\documentclass[fleqn,10pt]{wlscirep}
\usepackage[utf8]{inputenc}
\usepackage[T1]{fontenc}
\usepackage{amssymb}
\usepackage[ruled,vlined]{algorithm2e}
\usepackage{xcolor}
\usepackage{graphicx}
\usepackage{listings}
\usepackage{subcaption}
\usepackage{matlab-prettifier}
\usepackage[numbers]{natbib} 

\definecolor{cgreen}{rgb}{0,0.6,0}
\definecolor{cgray}{rgb}{0.5,0.5,0.5}
\definecolor{cpurple}{rgb}{0.58,0,0.82}
\definecolor{cwhite}{rgb}{1,1,1}

\title{Batch Normalization in Cytometry Data by kNN-Graph Preservation}

\author[1]{Muhammad S. Battikh\thanks{msbatikh@nmsu.edu}}
\author[2,3,4]{Artem Lensky\thanks{a.lenskiy@unsw.edu.au}}
\affil[1]{New Mexico State University, NM, USA}
\affil[2]{School of Engineering and Technology, The University of New South Wales, ACT, Australia}
\affil[3]{School of Biomedical Engineering, Faculty of Engineering, The University of Sydney, NSW, Australia}
\affil[3]{School of Computing, Australian National University, ACT, Australia}

\begin{abstract}
Batch effects in high-dimensional Cytometry by Time-of-Flight (CyTOF) data pose  a challenge for comparative analysis across different experimental conditions or time points. Traditional batch normalization methods may fail to preserve the complex topological structures inherent in cellular populations. In this paper, we present a residual neural network-based method for point set registration specifically tailored to address batch normalization in CyTOF data while preserving the topological structure of cellular populations. By viewing the alignment problem as the movement of cells sampled from a target distribution along a regularized displacement vector field, similar to coherent point drift (CPD), our approach introduces a Jacobian-based cost function and geometry-aware statistical distances to ensure local topology preservation. We provide justification for the k-Nearest Neighbour (kNN) graph preservation of the target data when the Jacobian cost is applied, which is crucial for maintaining biological relationships between cells. Furthermore, we introduce a stochastic approximation for high-dimensional registration, making alignment feasible for the high-dimensional space of CyTOF data. Our method is demonstrated on high-dimensional CyTOF dataset, effectively aligning distributions of cells while preserving the kNN-graph structure. This enables accurate batch normalization, facilitating reliable comparative analysis in biomedical research. 
\end{abstract}
\begin{document}

\flushbottom
\maketitle
\thispagestyle{empty}

\section*{Introduction}

Cytometry by Time-of-Flight (CyTOF) is a technology that allows simultaneous measurement of multiple biomarkers at the single-cell level, generating high-dimensional data essential for understanding complex biological systems. However, batch effects—systematic non-biological variations arising from differences in experimental conditions, instrumentation, or sample processing—pose significant challenges for the analysis and interpretation of CyTOF data~\cite{finck2013normalization, shaham2017removal}. These batch effects can obscure true biological differences and lead to incorrect conclusions, making effective batch normalization a critical step in CyTOF data analysis.

Traditional batch normalization methods often rely on aligning marginal distributions or applying linear transformations~\cite{ge2014non, hirose2020bayesian}. These methods may fail to capture the complex, non-linear relationships inherent in high-dimensional biological data and may not preserve the local topological structures and biological relationships between cells, such as cellular hierarchies and differentiation pathways~\cite{stoeckius2017simultaneous, peterson2017multiplexed}.

In this paper, we introduce a novel residual neural network-based method for point set registration specifically designed for batch normalization in CyTOF data. Our approach makes several key contributions:

\begin{itemize}
    \item A residual mapping that incorporates k-Nearest Neighbour (kNN) graph preservation to ensure that the local topological structure of the data is maintained during alignment. This is crucial for preserving biological relationships between cells, which are essential for downstream analyses.

    \item We introduce a Jacobian-based cost function that enforces the orthogonality of the Jacobian matrix of the transformation. This mathematical formulation provides a theoretical justification for kNN graph preservation and ensures that the local geometry of the data is preserved.

    \item Recognizing the computational challenges posed by the high dimensionality of CyTOF data, we implement Hutchinson's estimator to approximate the Jacobian regularization term efficiently. This makes our method computationally feasible for practical applications involving high-dimensional datasets.

    \item We apply our method to real-world CyTOF datasets, demonstrating its effectiveness in correcting batch effects while preserving biological integrity. 

\end{itemize}

By viewing the alignment problem as the movement of cells sampled from a target distribution along a regularized displacement vector field, similar to the coherent point drift (CPD) approach~\cite{myronenko2010point}, our method overcome limitations of existing techniques. Unlike CPD, which relies on a global smoothness constraint that can be too restrictive for locally rigid but globally non-rigid deformations common in biological data, our method employs a local regularization that better suits the complex deformations in CyTOF data. Our approach facilitates reliable comparative analyses and downstream biological interpretations, addressing a critical need in biomedical research involving high-dimensional single-cell data.

\begin{figure*}
\centering
         \centering
         \includegraphics[width=\textwidth, trim={2.5cm, 12cm, 1.5cm, 3.5cm}, clip]{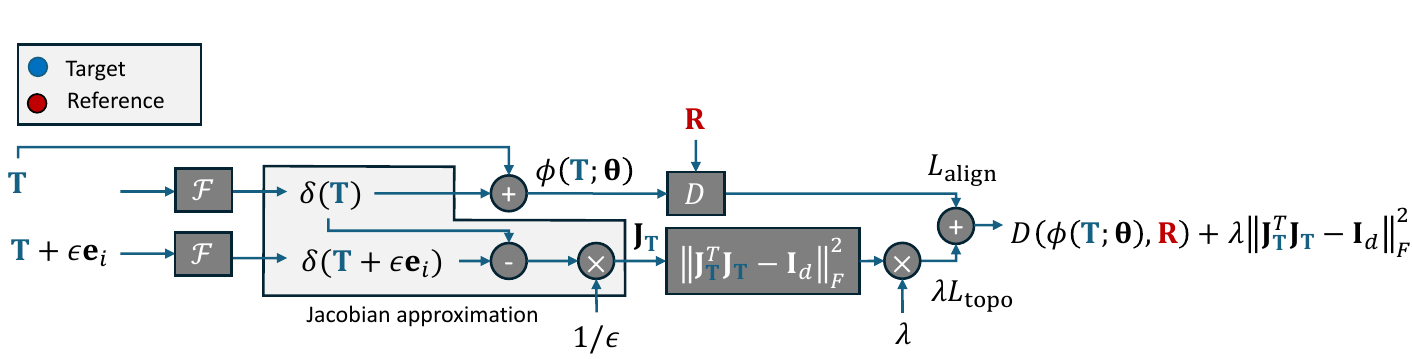}
         \caption{Training pipeline for the proposed alignment.}
         \label{fig:model}
\end{figure*}

\section*{Methods}

\subsection*{Overview}

The primary goal of our method is to correct batch effects in high-dimensional CyTOF data while preserving the biological relationships between cells. Batch effects can introduce systematic non-biological variations that obscure true biological differences, making it challenging to compare data across different batches or experimental conditions. Our approach aligns distributions of cells from different batches by learning a transformation that maps the target batch onto the reference batch, ensuring that the local topological structure (e.g., cellular hierarchies and differentiation pathways) is maintained.

\subsection*{Alignment Framework}

We model the batch normalization problem as a point set registration task, where each cell is represented as a point in a high-dimensional space defined by the measured biomarkers. Given two sets of cells, $\mathbf{R} = \{\mathbf{x}_1, \mathbf{x}_2, \dots, \mathbf{x}_n\}$ from the reference batch and $\mathbf{T} = \{\mathbf{y}_1, \mathbf{y}_2, \dots, \mathbf{y}_m\}$ from the target batch, we seek a transformation $\phi$ that aligns $\mathbf{T}$ to $\mathbf{R}$.

\subsection*{Residual Neural Network Transformation}

We parameterize the transformation $\phi$ using a residual neural network (ResNet) with identity blocks~\cite{he2016deep}, defined as:
\begin{equation}
    \phi(\mathbf{y}; \theta) = \mathbf{y} + \delta(\mathbf{y}; \theta),
\end{equation}
where $\mathcal{F}=\delta(\mathbf{y}; \theta)$ is a multilayer perceptron (MLP) representing the displacement field, and $\theta$ denotes the network parameters. This architecture naturally biases the transformation towards the identity mapping, encouraging minimal and smooth adjustments that correct batch effects without introducing non-biological distortions. A visualization of the proposed model and the training pipeline is shown in figure \ref{fig:model}.

\subsection*{Optimization Objective}

The overall loss function combines the alignment loss and the topology preservation regularization:

\begin{equation}
    \mathit{L}(\theta) = \mathit{L}_{\text{align}}(\theta) + \lambda \mathit{L}_{\text{topo}},
\end{equation}

where $\lambda$ is a hyperparameter controlling the trade-off between alignment accuracy and topology preservation. We optimize this loss function with respect to the network parameters $\theta$ using stochastic gradient descent.

\subsection*{Alignment Loss Function}

We employ geometry-aware statistical distances to measure the discrepancy between the transformed target distribution and the reference distribution. Specifically, we use the Sinkhorn divergence~\cite{feydy2019interpolating}, which interpolates between the Wasserstein distance and the Maximum Mean Discrepancy (MMD):

\begin{equation}
    \mathit{L}_{\text{align}}(\theta) = S_{\epsilon}(\alpha, \beta) = \text{OT}_{\epsilon}(\alpha, \beta) - \frac{1}{2} \left( \text{OT}_{\epsilon}(\alpha, \alpha) + \text{OT}_{\epsilon}(\beta, \beta) \right),
\end{equation}

where $\alpha$ and $\beta$ are empirical measures over $\mathbf{R}$ and $\phi(\mathbf{T}; \theta)$, respectively, and $\text{OT}_{\epsilon}$ denotes the entropic regularized optimal transport. This loss function captures both the global and local differences between the distributions.

\subsection*{Topology Preservation via Jacobian Regularization}

\begin{figure}
    \centering
    \includegraphics[width=\textwidth, trim={0cm, 6cm, 0cm, 5cm}, clip]{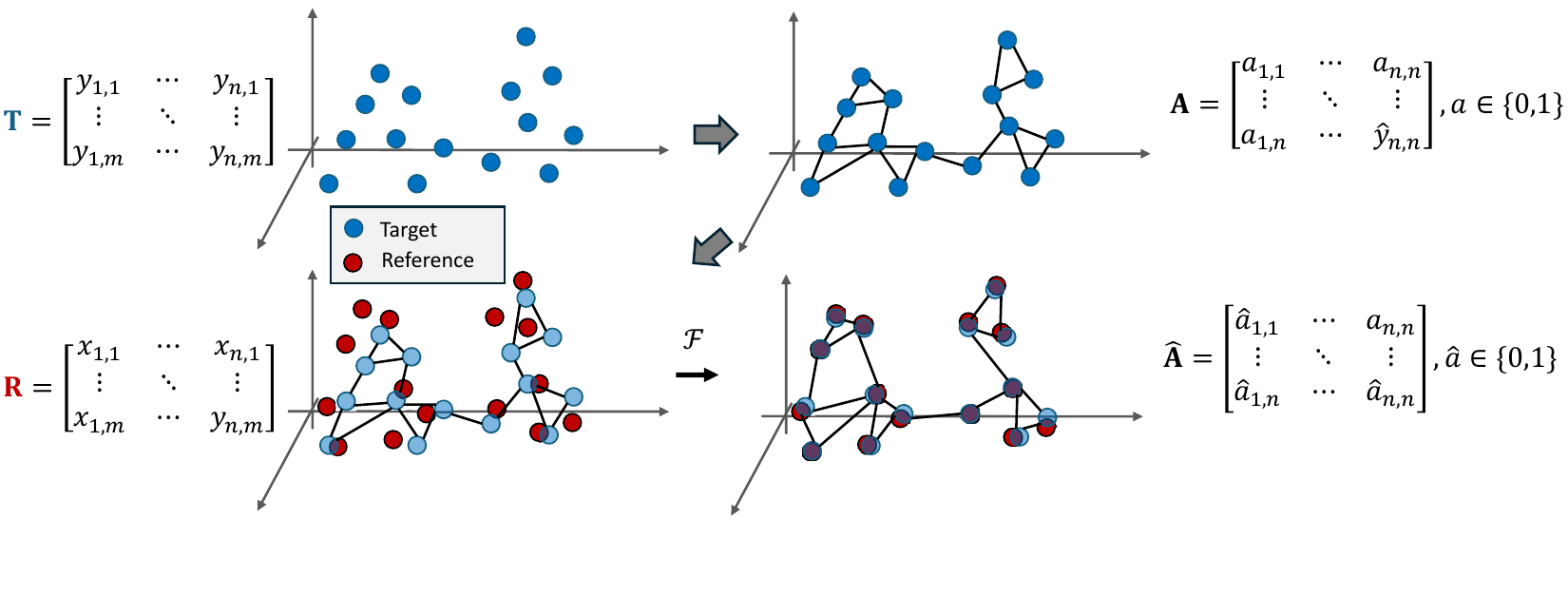}
    \caption{Conceptual illustration of topology preserving alignment. The proposed algorithm does not construct the kNN graph.}
    \label{fig:knn-illustration}
\end{figure}

To ensure that the local topological structure of the cellular populations is preserved during alignment, we introduce a regularization term based on the Jacobian matrix $\mathbf{J}_{\mathbf{y}}$ of the transformation at each point $\mathbf{y}$:
\begin{equation}
    \mathit{L}_{\text{topo}} = \frac{1}{m} \sum_{\mathbf{y} \in \mathbf{T}} \left\| \mathbf{J}_{\mathbf{y}}^\top \mathbf{J}_{\mathbf{y}} - \mathbf{I}_d \right\|_F^2,
\end{equation}
where $\mathbf{I}_d$ is the $d \times d$ identity matrix, and $\left\| \cdot \right\|_F$ denotes the Frobenius norm. This regularization encourages the transformation to be locally rigid (i.e., approximately orthogonal), which helps preserve the k-Nearest Neighbour (kNN) graph structure of the data, maintaining the biological relationships between cells (Fig. \ref{fig:knn-illustration}). This process could be computed efficiently in a few lines of code as indicated in algorithm \ref{algo:1}.

\subsection*{Computational Considerations for High-Dimensional Data}
Using finite difference to compute the Jacobian for low-dimensional point clouds is efficient, however, the computational cost increases linearly with the dimension of the data.  Thus, an approximate estimate with the constant computational cost is introduced. 

Given a vector-valued function $\mathcal{F}$, and a sample $\mathbf{x}$, we would like to minimize the following: 
\begin{equation}
   \mathit{L}_{\text{topo}}(\mathcal{F})= \vert \mathbf{J}^\top \mathbf{J} \circ (1 - \mathbf{I}) \vert_2 =  \sum_{i \neq j} \frac{\partial \mathcal{F}_i}{\partial x_j} \frac{\partial \mathcal{F}_j}{ \partial x_i} 
\end{equation}

Following~\cite{wei2021orthogonal, hutchinson1989stochastic}, the Hutchinson’s estimator of $\mathit{L}_{\mathbf{J}}(F)$ can be approximated as such: 
\begin{equation}
\mathit{L}_{\text{topo}}(\mathit{L}) = \text{Var}_{\mathbf{r}} (\mathbf{r}^\top_{\epsilon} (\mathbf{J}^\top \mathbf{J})\mathbf{r}_\epsilon )= \text{Var}_\mathbf{r}((\mathbf{J}\mathbf{r}_\epsilon)^\top (\mathbf{J}\mathbf{r}_\epsilon))    
\end{equation}

where $r_\epsilon$ denotes a scaled Rademacher vector (each entry is either $-\epsilon$ or $+\epsilon$ with equal probability) where $\epsilon>0$ is a hyperparameter that controls the granularity of the first directional derivative estimate and $\text{Var}_r$  is the variance. It is worth noting that this does not guarantee orthonormality, only orthogonality. In practice, however, we find that such an estimator produces comparable results to the standard finite difference method and could be efficiently implemented in Matlab as shown in algorithm \ref{algo:hutchinsons_estimator}. 

The function $\texttt{orth\_jacobian\_fin\_diff}$ can be used when the dimensionality is manageable ($<10$), while \\$\texttt{stochastic\_orth\_jacobian}$ should be used for computational efficiency when the number of dimensions is high.

\subsection*{Dataset and Preprocessing}

\subsubsection*{Chui-Rangarajan synthesized dataset}

Before testing the batch normalization on the flow cytometry dataset ~\cite{shaham2017removal}, we evaluate the method on the synthesized dataset, introduced by Chui-Rangarajan  in~\cite{chui2003new,zheng2006robust,ma2013robust}that is comprised of two shapes; a fish shape, and a Chinese character shape. Each shape is subjected to 5 increasing levels of deformations using an RBF kernel, and each deformation contains 100 different samples. The samples are generated using different RBF coefficients which are sampled from a zero-mean normal distribution with standard deviation $\sigma$, whereby increasing $\sigma$ leads to generally larger deformation. 
\subsubsection*{Yale New Haven Hospital CyTOF dataset}
The CyTOF dataset used in our experiments was curated by the Yale New Haven Hospital. It comprises measurements from two patients under two different conditions, collected on two separate days. In total, there are eight samples, each with 25 biomarkers per cell, representing separate dimensions: \textit{CD45}, \textit{CD19}, \textit{CD127}, \textit{CD4}, \textit{CD8a}, \textit{CD20}, \textit{CD25}, \textit{CD278}, \textit{TNFa}, \textit{Tim3}, \textit{CD27}, \textit{CD14}, \textit{CCR7}, \textit{CD28}, \textit{CD152}, \textit{FOXP3}, \textit{CD45RO}, \textit{INFg}, \textit{CD223}, \textit{GzB}, \textit{CD3}, \textit{CD274}, \textit{HLADR}, \textit{PD1}, and \textit{CD11b}. The number of cells per sample ranges from 1,800 to 5,000.

We split the data such that samples collected on Day 1 serve as the target datasets, and samples collected on Day 2 serve as the reference datasets, resulting in four alignment experiments. This setup allows us to evaluate the effectiveness of our method in correcting batch effects across different days.

We followed the exact preprocessing steps described in~\citep{shaham2017removal}. To adjust the dynamic range of the samples, a standard preprocessing step of applying a log transformation was performed~\cite{finck2013normalization}. Additionally, CyTOF data typically contains a large number of zero values (approximately 40\%) due to instrumental instability, which are not considered biological signals. To address this, we used a denoising autoencoder (DAE) to remove these zero values~\cite{vincent2008extracting}.

The encoder of the DAE comprises two fully connected layers with ReLU activation functions, and the decoder is a single linear layer without an activation function. All layers have the same number of neurons as the dimensionality of the data (25 neurons). During training, each cell is multiplied by an independent Bernoulli random vector with a probability of 0.8 (dropout), and the DAE is trained to reconstruct the original cell using a mean squared error. The DAE is optimized using RMSprop with weight decay regularization. After training, the zero values in both reference and target datasets are removed using the trained DAE. Finally, each feature in both target and reference samples is independently standardized to have zero mean and unit variance.

% \subsection*{Experimental Setup}

% We conducted four alignment experiments corresponding to the different patient samples and conditions. For the proposed model, we set the hyperparameters as follows: $\epsilon = 0.05$, $\lambda = 0.1$, $\sigma = 0.04$, $k = 5$ for Hutchinson's estimator, and the number of hidden units in the MLP to 50. We started with a relatively high learning rate of 0.01 for the Adam optimizer and used a reduce-on-plateau scheduler with a reduction factor of 0.7 and a minimum learning rate of $5 \times 10^{-5}$.

% We trained five models with different values of $\lambda$ ranging from 0 to 1 to observe the effect of the regularization parameter on alignment performance and topology preservation.

\subsection*{Training Procedure}

We trained the residual neural network using mini-batch stochastic gradient descent as outlined in Algorithm~\ref{algo:training}. In each iteration, we sampled mini-batches from the reference and target datasets, computed the transformed target points, and evaluated the alignment and topology preservation losses. The network parameters were updated to minimize the total loss.

\begin{algorithm}[t!]
\SetAlgoLined
\begin{lstlisting}[style=Matlab-editor, basicstyle=\footnotesize]
function loss = orth_jacobian_fin_diff(G, z, epsilon)
    % Input G: dlnetwork object to compute the Jacobian Penalty for.
    % Input z: (d, batchsize) Input to G that the Jacobian is computed w.r.t.
    % Input epsilon: (default 0.01) Step size for finite difference
    % Output: mean(|J_X^T J_X - I_d|)
    
    if nargin < 3
        epsilon = 0.01;
    end
    
    [d, batchsize] = size(z);
    Gz = predict(G, z); % Forward pass on original input
    % Repeat output for each dimension perturbation
    Gz_rep = reshape(repmat(Gz, 1, d), [d*batchsize, d]);
    % Create identity matrices for perturbations
    I = dlarray(repmat(eye(d, 'single'), 1, 1, batchsize));
    % Expand input for vectorized perturbation
    z_expanded = reshape(repmat(z, 1, d), [d, d*batchsize])';
    % Apply perturbations: z + epsilon * e_i for each dimension i
    zdz = z_expanded + epsilon * reshape(I, [d*batchsize, d]);
    out = predict(G, dlarray(zdz, 'CB')); % Forward pass on all perturbed inputs
    jac = ((out - Gz_rep) / epsilon); % Compute Jacobian using (f(z+h) - f(z))/h
    % Reshape to [d, d, batchsize] where jac(:,:,i) is Jacobian for sample i
    jac = reshape(jac, [d, d, batchsize]);
    JtJ = pagemtimes(pagectranspose(jac), jac); % Compute J^T * J for each sample
    I_batch = repmat(eye(d, 'single'), 1, 1, batchsize); % Create identity matrices
    bloss = JtJ - I_batch; % Compute deviation from orthogonality
    loss = mean(abs(bloss), 'all'); % Return mean absolute deviation
end
\end{lstlisting}
    \caption{MATLAB code for Jacobian deviation from Orthogonality at data points $z$ using finite difference method.}
\label{algo:1}
\end{algorithm}

\begin{algorithm}[t!]
\SetAlgoLined
\begin{lstlisting}[style=Matlab-editor, basicstyle=\footnotesize]
function loss = stochastic_orth_jacobian(G, z, k, eps)
    % Input G: dlnetwork object to compute the Jacobian Penalty for.
    % Input z: (d, batchsize) Input to G that the Jacobian is computed w.r.t.
    % Input k: number of directions to sample (default 5)
    % Input eps: (default 0.01) Step size for finite difference
    % Output: mean(|J_X^T J_X - I_d|)
    
    if nargin < 3
        k = 5;
    end
    if nargin < 4
        eps = 0.01;
    end
    
    [d, batchsize] = size(z);
    Gz = predict(G, z); % Forward pass on original input
    % Generate Rademacher random vectors: {-1, +1}^{k x d x batchsize}
    r = randi([0, 1], k, d, batchsize, 'single');
    r(r == 0) = -1;
    % Scale by epsilon
    vs = eps * r;
    % Compute stochastic finite differences
    diffs = zeros(k, d, batchsize, 'single');
    for i = 1:k
        % Apply perturbation v_i to all samples
        v_i = dlarray(squeeze(vs(i,:,:)), 'CB');
        perturbed_out = predict(G, z + v_i);
        diffs(i,:,:) = perturbed_out - Gz;
    end
    
    % Convert to dlarray and normalize by epsilon
    sfd = dlarray(diffs) / eps;
    % Compute variance across random directions and take maximum
    loss = max(var(sfd, 0, 1), [], 'all');
end
\end{lstlisting}
    \caption{MATLAB code for Hutchinson approximation for Jacobian off-diagonal elements at data points $z$.}
\label{algo:hutchinsons_estimator}
\end{algorithm}

% \begin{algorithm}[t!]
% \caption{Training Procedure for Batch Normalization in CyTOF Data}
% \SetAlgoLined
% \DontPrintSemicolon
% \KwIn{Reference dataset $\mathbf{R}$, target dataset $\mathbf{T}$, hyperparameters $\lambda$, $\epsilon$, $\sigma$, batch size $b$}
% \KwOut{Trained network parameters $\theta$}
% Initialize network parameters $\theta$\;
% \While{not converged}{
%     Sample mini-batches $\mathbf{X} \subset \mathbf{R}$ and $\mathbf{Y} \subset \mathbf{T}$\;
%     Compute transformed target points $\phi(\mathbf{Y}; \theta)$\;
%     Compute alignment loss $\mathit{L}_{\text{align}}(\theta)$ using Sinkhorn divergence\;
%     Approximate Jacobian regularization $\mathit{L}_{\text{topo}}$ using Hutchinson's estimator\;
%     Compute total loss $\mathit{L}(\theta) = \mathit{L}_{\text{align}}(\theta) + \lambda \mathit{L}_{\text{topo}}$\;
%     Update network parameters $\theta$ using gradient descent\;
% }
% \label{algo:training}
% \end{algorithm}

\begin{algorithm}[t!]
\caption{Training Procedure for Batch Normalization in CyTOF Data (MATLAB)}
\SetAlgoLined
\DontPrintSemicolon
\begin{lstlisting}[style=Matlab-editor, basicstyle=\footnotesize]
function net = trainBatchNormalization(R, T, lam, eps, sig, batchSize, maxIter)
    % Input R: Reference dataset (d, n_ref)
    % Input T: Target dataset (d, n_target) 
    % Input lam: Topology preservation weight
    % Input eps: Finite difference step size
    % Input sig: Sinkhorn divergence bandwidth
    % Input batchSize: Mini-batch size
    % Input maxIter: Maximum training iterations
    % Output net: Trained dlnetwork object
    
    % Initialize dlnetwork with residual architecture
    inputDim = size(R, 1);

    net = dlnetwork(layerGraph([featureInputLayer(inputDim), ...
              fullyConnectedLayer(50), ...
              leakyReluLayer(0.05), ...
              fullyConnectedLayer(50), ...
              fullyConnectedLayer(inputDim)]));
    
    % Initialize Adam optimizer state
    avgGrad = [];
    avgSqGrad = [];
    lr = 0.01;
    
    for iter = 1:maxIter
        % Sample mini-batches from datasets
        X = dlarray(datasample(R, batchSize, 2), 'CB');
        Y = dlarray(datasample(T, batchSize, 2), 'CB');
        % Compute gradients using automatic differentiation
        [gradients, loss, alignLoss, topoLoss] = dlfeval(@modelGradients, net, Y, X, eps, lamb, sig);
        % Update network parameters using Adam
        [net, avgGrad, avgSqGrad] = adamupdate(net, gradients, avgGrad, avgSqGrad, iter, lr);
    end
end

function [gradients, totalLoss, alignLoss, topoLoss] = modelGradients(net, Y, X, eps, lam, sig)
    % Transform target points: phi(Y; theta)
    transformedY = predict(net, Y) + Y;  % Residual connection
    alignLoss = computeSinkhornLoss(transformedY, X, sig); % Compute alignment loss using Sinkhorn divergence
    topoLoss = orth_jacobian_fin_diff(net, Y, eps); % Approx. Jacobian regularization using finite diff.
    totalLoss = alignLoss + lam * topoLoss; % Total loss: L(theta) = L_align(theta) + lam * L_topo
    gradients = dlgradient(totalLoss, net.Learnables); % Compute gradients
end
\end{lstlisting}
\label{algo:training}
\end{algorithm}

\section*{Results}
\subsubsection*{Chui-Rangarajan synthesized dataset}
\begin{figure}
     \centering
     \begin{subfigure}[b]{0.145\textwidth}
         \centering
         \includegraphics[width=\textwidth]{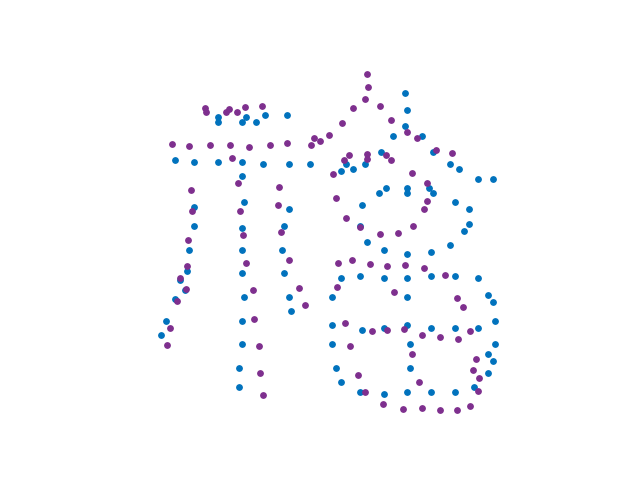}
     \end{subfigure}
    \hfill
     \begin{subfigure}[b]{0.145\textwidth}
         \centering
         \includegraphics[width=\textwidth]{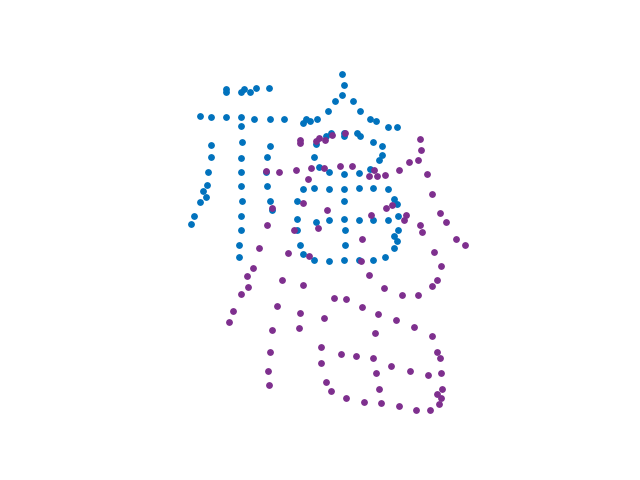}
     \end{subfigure}
    \hfill
     \begin{subfigure}[b]{0.145\textwidth}
         \centering
         \includegraphics[width=\textwidth]{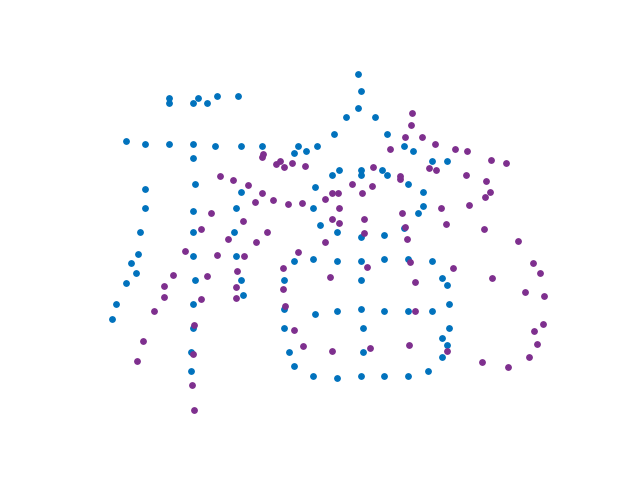}
     \end{subfigure}
    \hfill
     \begin{subfigure}[b]{0.19\textwidth}
         \centering
         \includegraphics[width=\textwidth]{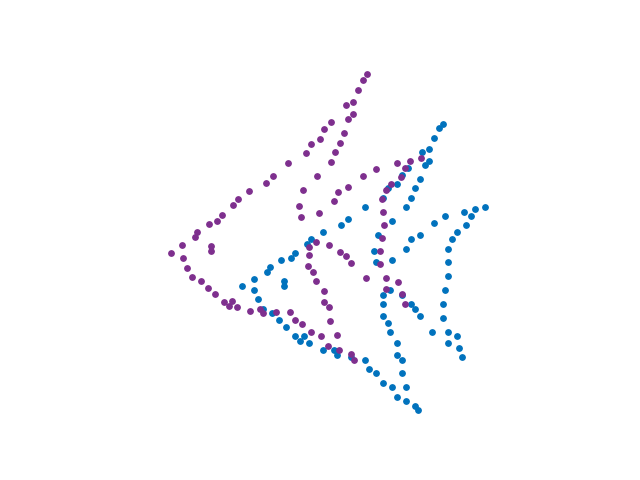}
     \end{subfigure}
    \hfill
     \begin{subfigure}[b]{0.145\textwidth}
         \centering
         \includegraphics[width=\textwidth]{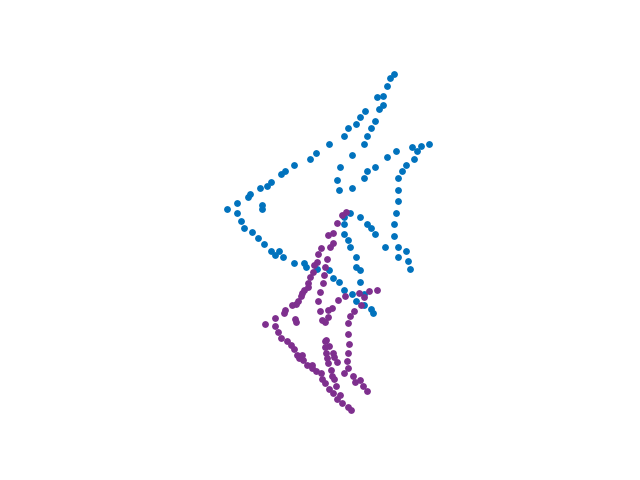}
     \end{subfigure}
    \hfill
     \begin{subfigure}[b]{0.145\textwidth}
         \centering
         \includegraphics[width=\textwidth]{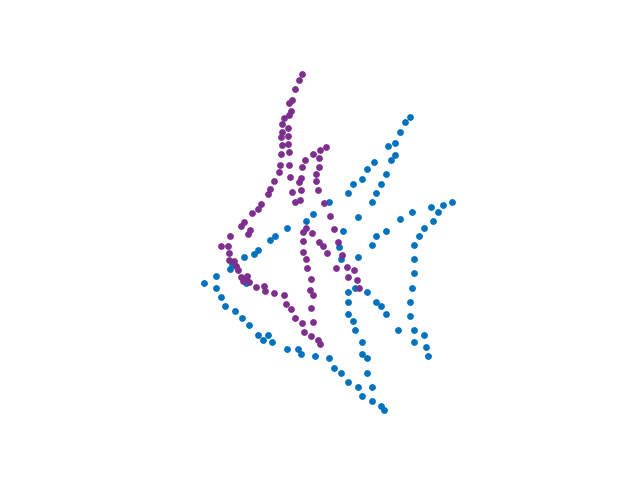}
     \end{subfigure}
\\
     \begin{subfigure}[b]{0.145\textwidth}
         \centering
         \includegraphics[width=\textwidth]{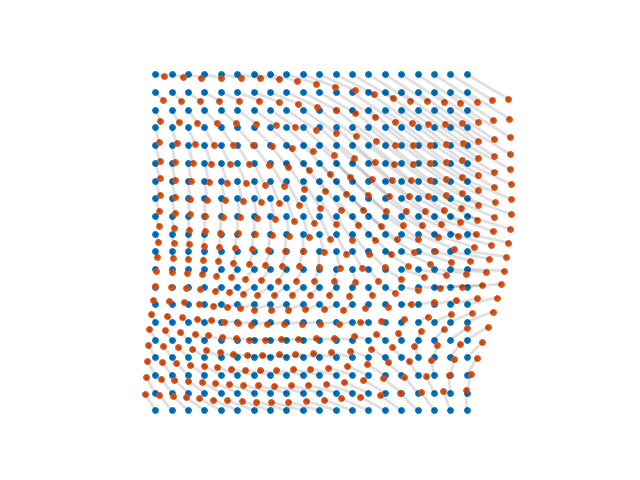}
     \end{subfigure}
    \hfill
     \begin{subfigure}[b]{0.145\textwidth}
         \centering
         \includegraphics[width=\textwidth]{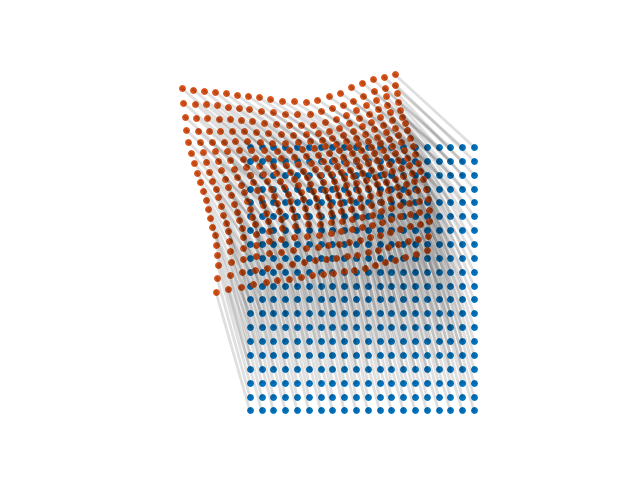}
     \end{subfigure}
    \hfill
     \begin{subfigure}[b]{0.19\textwidth}
         \centering
         \includegraphics[width=\textwidth]{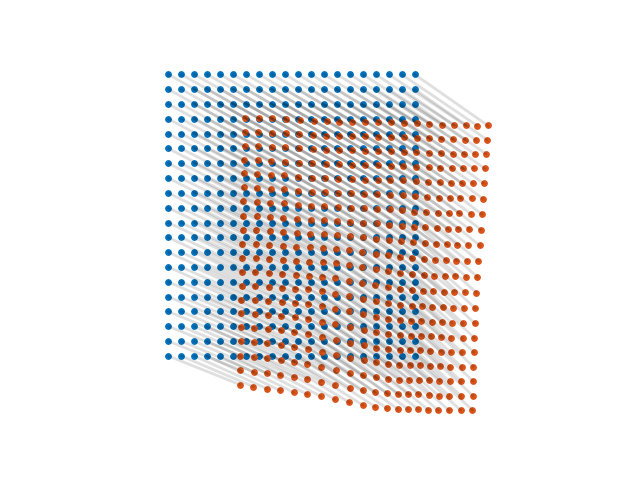}
     \end{subfigure}
    \hfill
     \begin{subfigure}[b]{0.145\textwidth}
         \centering
         \includegraphics[width=\textwidth]{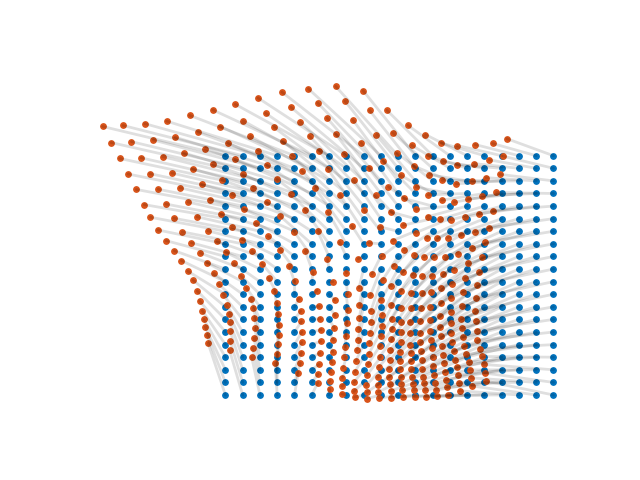}
     \end{subfigure}
    \hfill
     \begin{subfigure}[b]{0.145\textwidth}
         \centering
         \includegraphics[width=\textwidth]{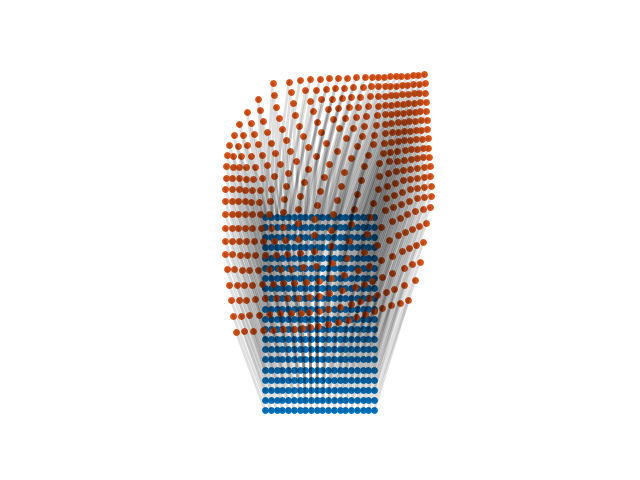}
     \end{subfigure}
    \hfill
     \begin{subfigure}[b]{0.145\textwidth}
         \centering
         \includegraphics[width=\textwidth]{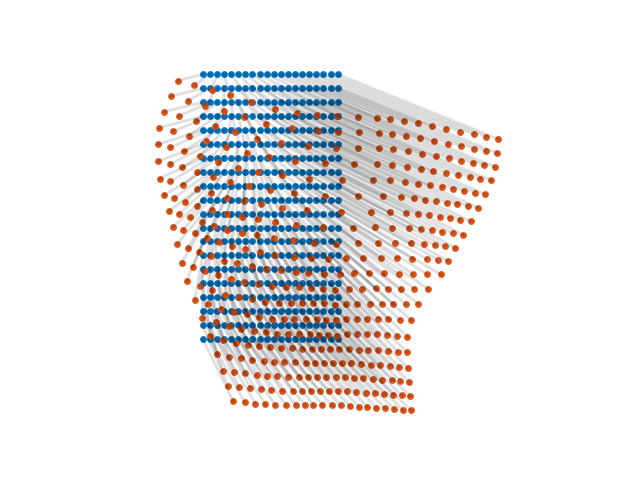}
     \end{subfigure}
\\
     \begin{subfigure}[b]{0.145\textwidth}
         \centering
         \includegraphics[width=\textwidth]{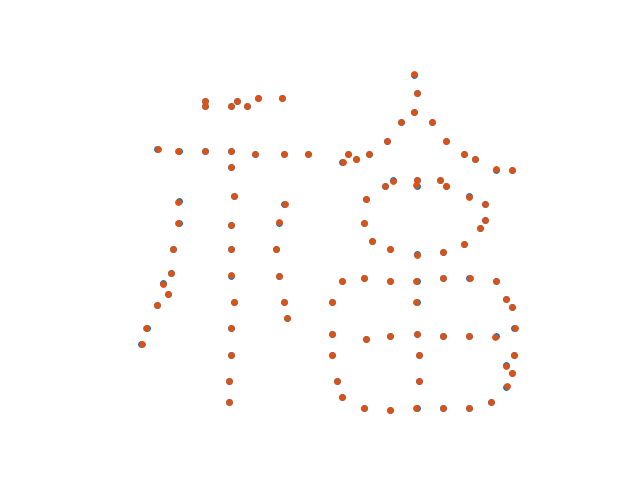}
     \end{subfigure}
    \hfill
     \begin{subfigure}[b]{0.145\textwidth}
         \centering
         \includegraphics[width=\textwidth]{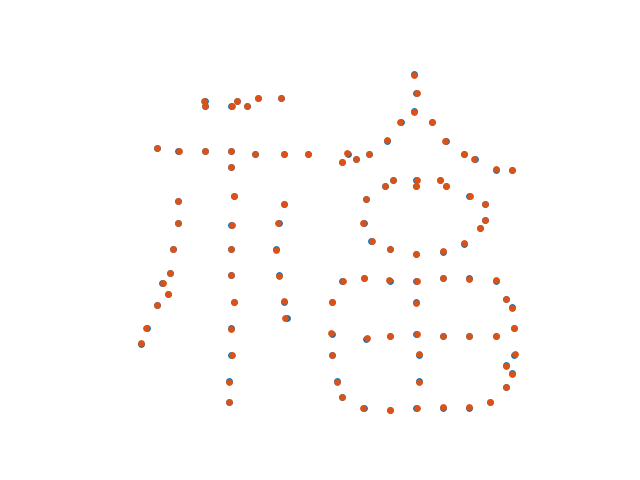}
     \end{subfigure}
    \hfill
     \begin{subfigure}[b]{0.19\textwidth}
         \centering
         \includegraphics[width=\textwidth]{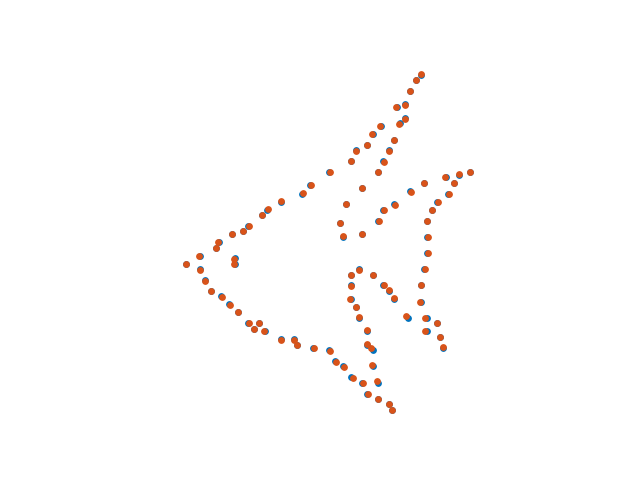}
     \end{subfigure}
    \hfill
     \begin{subfigure}[b]{0.145\textwidth}
         \centering
         \includegraphics[width=\textwidth]{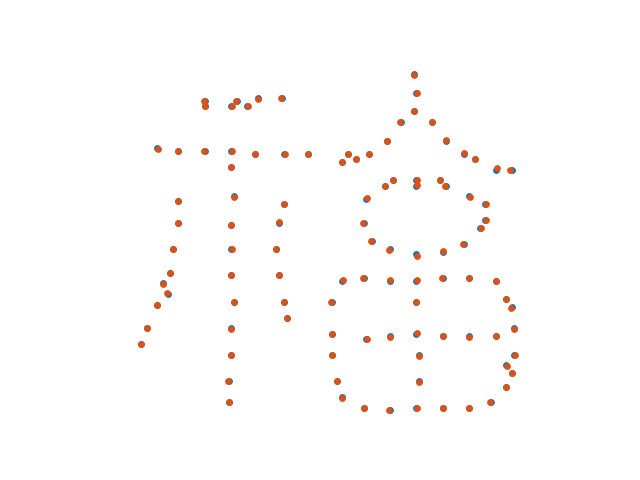}
     \end{subfigure}
    \hfill
     \begin{subfigure}[b]{0.145\textwidth}
         \centering
         \includegraphics[width=\textwidth]{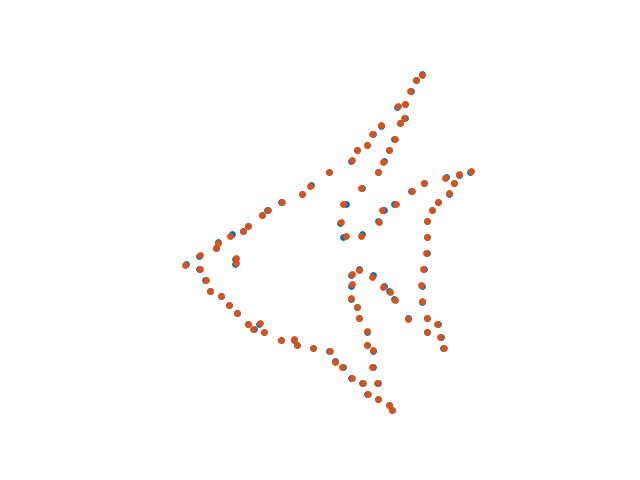}
     \end{subfigure}
    \hfill
     \begin{subfigure}[b]{0.145\textwidth}
         \centering
         \includegraphics[width=\textwidth]{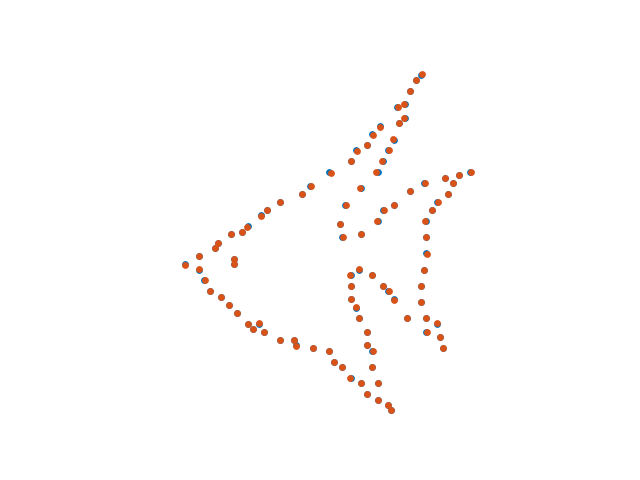}
     \end{subfigure}
    \caption{Several examples of deformations of the Chinese character and Fish: Top row represents original and deformed sets, Mid row represents the vector field, and Bottom row is the final alignment.}
    \label{fig:the_chinese_character_and_fish_example}
\end{figure}

We use the root-mean-squared error (RMSE) between the transformed data $\hat{y_i}$ and the ground truth $y_i$ available from the Chui-Rangarajan synthesized dataset:
$\texttt{RMSE} = \sqrt{\frac{1}{m} \sum_{i=0}^{m}{(\hat{y_i} - y_i)^2}}$.
It is important to note that such ground-truth correspondence is absent during training time and is only available during test time. Figure \ref{fig:the_chinese_character_and_fish_example} show the initial point set distributions and their corresponding aligned versions for the Chinese character and the fish examples respectively. We also report results for our model, MM-Res\cite{shaham2017removal}, CPD~\cite{myronenko2010point}, TRS-RPM~\cite{chui2003new}, RPM-LNS~\cite{ma2013robust}, and GMMREG~\cite{jian2010robust} over 5 deformation levels and 100 samples per level. Table \ref{tab:evaluation_metrics} shows results for tested models on the Chinese character, and Fish datasets respectively. We notice that after a certain level of non-rigid deformation, MM-Res is unable to converge.
For the proposed model, we set $\epsilon=.005, \lambda =10^{-5}, \sigma = .001$ and number of hidden units $N = 50$. We start with a relatively high learning rate (0.01) for ADAM~\cite{kingma2014adam} optimizer and use a reduce-on-plateau scheduler with a reduction factor of  0.7 and minimum learning rate of $5 \times 10^{-5}$. Qualitatively, the grid-warp representations from the second column in figure \ref{fig:the_chinese_character_and_fish_example} indicate that our estimated transformations are, at least visually, "simple" and "coherent". Furthermore, to quantitatively assess neighborhood preservation we use the hamming loss $\mathit{L}_{H}$ to estimate the difference between the kNN graph before and after transformation:
$$\mathit{L}_{H} = \sum_{i=0}^{m}{\sum_{j=0}^{m}{I(\hat{a}_{i,j}^{k} \neq a_{i,j}^{k})}}$$
where $a_{i,j}^{k}$ is the $i$,$j$ element of the k-NN graph matrix $\mathbf{A}$ before transformation, $\hat{a}_{i,j}^{k}$ is an element of $\hat{\mathbf{A}}$ and represent the corresponding element after transformation, and $I$ is the indicator function. 
Table \ref{tab:hamming_loss} show the difference in neighborhood preservation between MM-Res and the proposed model for the Chinese character, and Fish datasets respectively for three different levels of deformations.

\begin{table}[h!]
\centering
\caption{Evaluation metrics for different methods across deformation levels}
\label{tab:evaluation_metrics}
\begin{tabular}{c| c c c c c c}
\toprule
\textbf{Deformation Level} & \textbf{Proposed Model} & \textbf{MM-Res} & \textbf{CPD} & \textbf{RPM-L2E} & \textbf{TPS-RPM} & \textbf{GMM-REG} \\
\midrule
\multicolumn{7}{c}{\textbf{Chinese Character}} \\
\midrule
0.02 & 0.005 & 0.006 & 0.005 & 0.005 & 0.006 & 0.007 \\
0.04 & \textbf{0.010} & 0.012 & 0.011 & 0.013 & 0.015 & 0.018 \\
0.06 & \textbf{0.020} & 0.022 & 0.023 & 0.027 & 0.030 & 0.035 \\
0.08 & \textbf{0.035} & 0.040 & 0.042 & 0.048 & 0.053 & 0.060 \\ \midrule
\multicolumn{7}{c}{\textbf{Fish}} \\
\midrule
0.02 & \textbf{0.004} & 0.005 & 0.005 & 0.006 & 0.006 & 0.007 \\
0.04 & \textbf{0.009} & 0.011 & 0.012 & 0.013 & 0.014 & 0.017 \\ 
0.06 & \textbf{0.018} & 0.021 & 0.023 & 0.026 & 0.028 & 0.032 \\ 
0.08 & \textbf{0.030} & 0.034 & 0.038 & 0.043 & 0.047 & 0.052 \\ \bottomrule
\end{tabular}
\end{table}

\begin{table}[h!]
\centering
\caption{Hamming loss for the proposed model and MM-Res at different deformation levels and values of \( k \). Top row corresponds to Fish shape and bottom row corresponds to Chinese character.}
\label{tab:hamming_loss}
\begin{tabular}{c | c c c c c c }
\toprule
\textbf{\( k \)} & \textbf{Level 1 (Fish)} & \textbf{Level 2 (Fish)} & \textbf{Level 3 (Fish)} & \textbf{Level 1 (Chinese)} & \textbf{Level 2 (Chinese)} & \textbf{Level 3 (Chinese)} \\ 
\bottomrule
\multicolumn{7}{ c }{\textbf{proposed model}} \\
\bottomrule
1 & 0.002 & 0.004 & 0.005 & 0.002 & 0.004 & 0.005 \\ 
5 & 0.007 & 0.015 & 0.020 & 0.007 & 0.015 & 0.021 \\ 
10 & 0.012 & 0.025 & 0.032 & 0.012 & 0.025 & 0.033 \\
\midrule
\multicolumn{7}{ c }{\textbf{MM-Res}} \\ 
\midrule
1 & 0.003 & 0.005 & 0.006 & 0.003 & 0.005 & 0.006 \\ 
5 & 0.008 & 0.016 & 0.021 & 0.008 & 0.016 & 0.022 \\
10 & 0.013 & 0.026 & 0.034 & 0.013 & 0.027 & 0.035 \\ 
\bottomrule
\end{tabular}
\end{table}

\subsubsection*{Yale New Haven Hospital CyTOF dataset}

\begin{figure}[h]
\centering
     \begin{subfigure}[b]{0.33\textwidth}
         \centering
         \includegraphics[width=\textwidth, trim={1cm, 6cm, 1cm, 5cm}, clip]{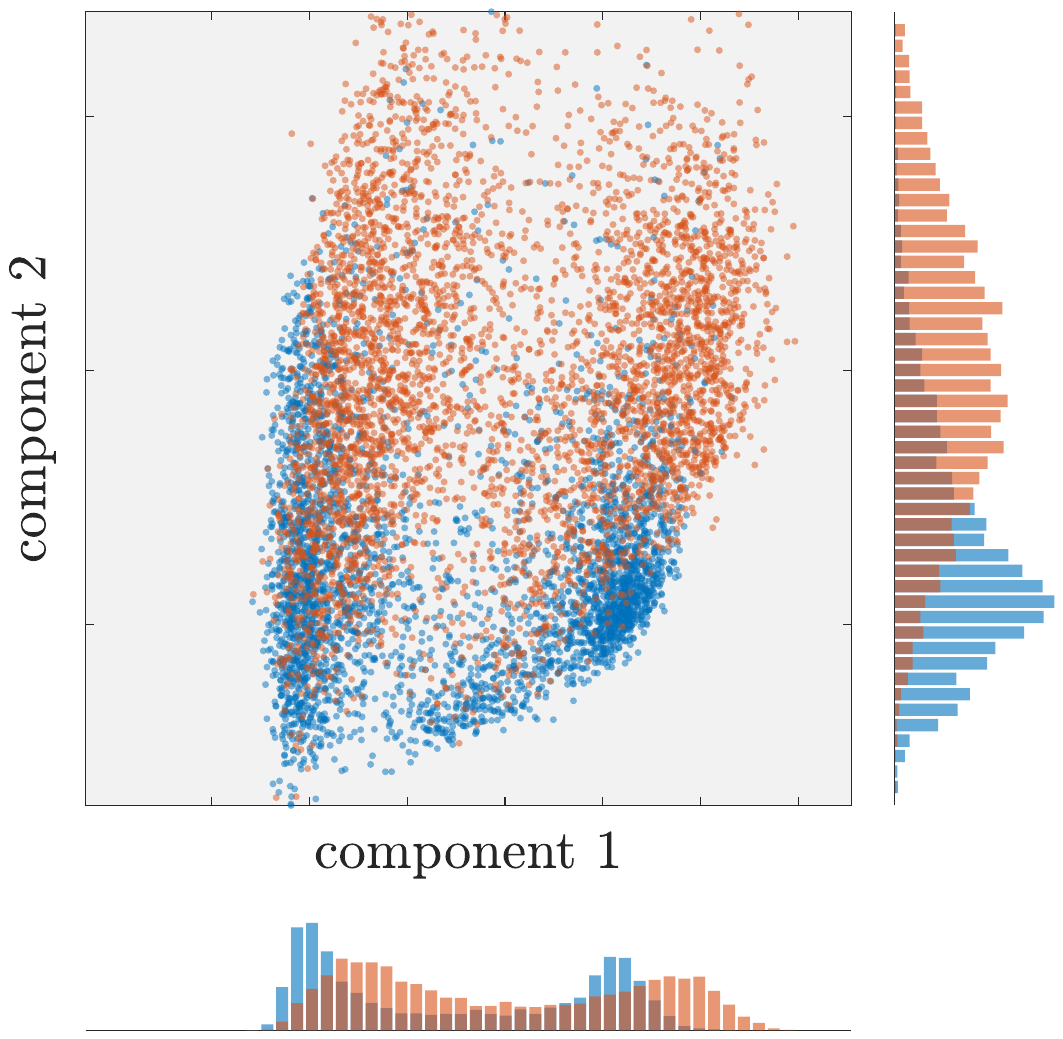}
         \caption{Unaligned Data}
     \end{subfigure}
     \begin{subfigure}[b]{0.33\textwidth}
         \centering
         \includegraphics[width=\textwidth, trim={1cm, 6cm, 1cm, 5cm}, clip]{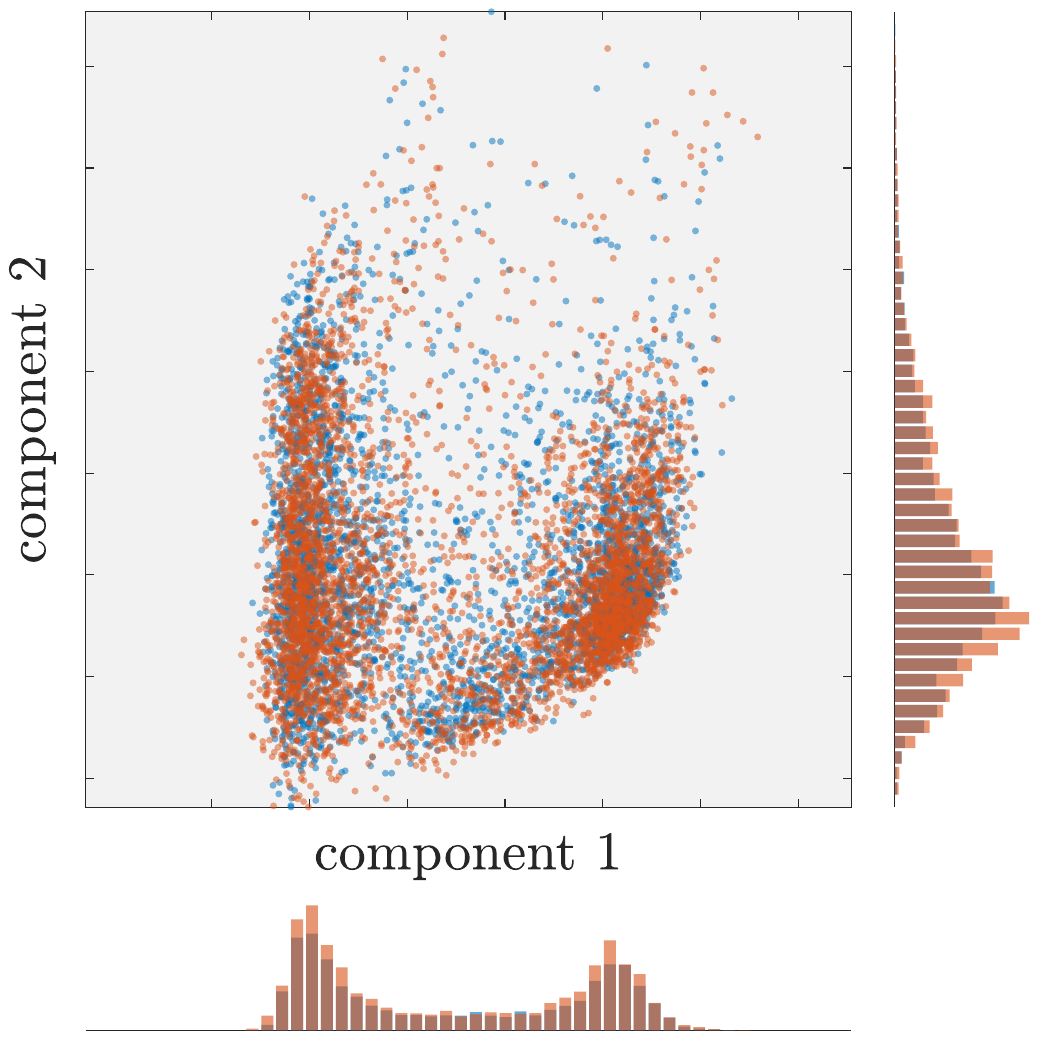}
         \caption{Aligned without Constraints ($\lambda = 0$)}
     \end{subfigure}
     \begin{subfigure}[b]{0.33\textwidth}
         \centering
         \includegraphics[width=\textwidth, trim={1cm, 6cm, 1cm, 5cm}, clip]{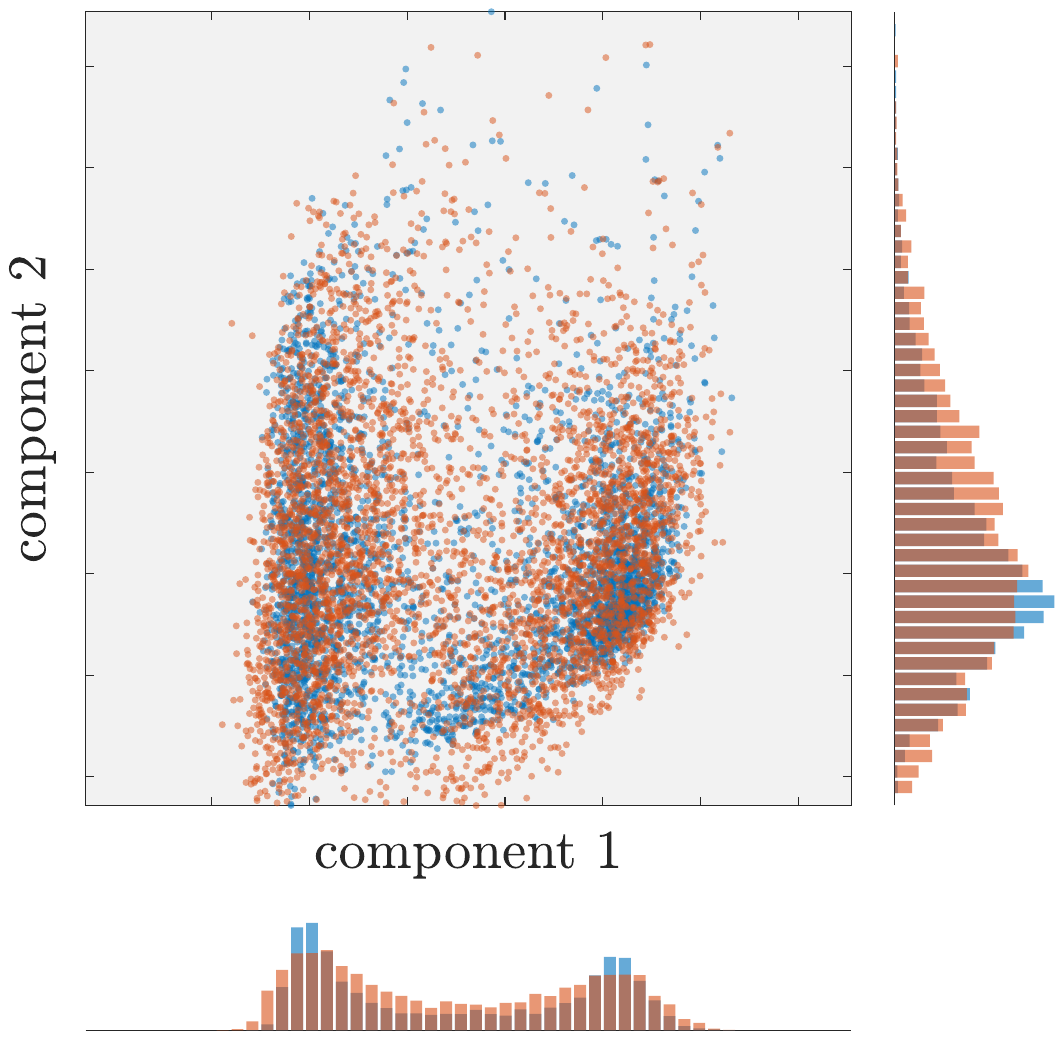}
         \caption{Aligned with Constraints ($\lambda = 1$)}
     \end{subfigure}
    \caption{Principal Component Analysis (PCA) plots of patient \#2 samples on Day 1 (target, red) and Day 2 (reference, blue). (a) Unaligned data shows batch effects. (b) Alignment without topology preservation ($\lambda = 0$) reduces batch effects but may distort local structures. (c) Alignment with topology preservation ($\lambda = 1$) maintains local structures while correcting batch effects.}
    \label{fig:person2_PCA_comparison}
\end{figure}

Our method effectively aligned the target samples to the reference samples. Figure~\ref{fig:person2_PCA_comparison} shows the first two principal components of data before and after alignment for patient \#2. The batch effects were substantially reduced, bringing the distributions of cells from different batches into closer agreement.

\begin{figure}[h]
\centering
     \begin{subfigure}[b]{0.33\textwidth}
         \centering
         \includegraphics[width=\textwidth, trim={1cm, 5.5cm, 1cm, 5.5cm}, clip]{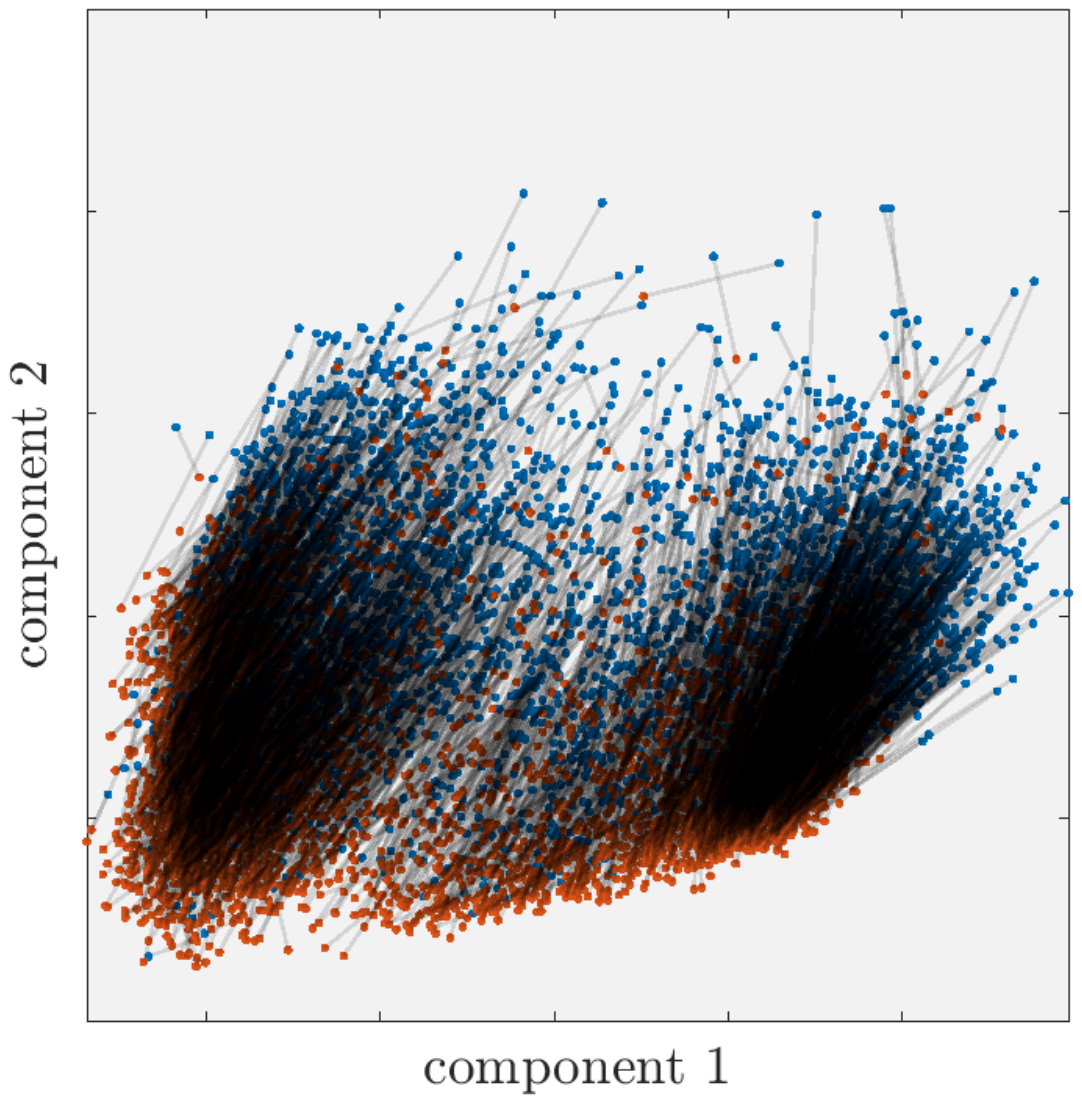}
         \caption{Alignment without Constraints ($\lambda = 0$)}
     \end{subfigure}
     \hfill
      \begin{subfigure}[b]{0.33\textwidth}
         \centering
         \includegraphics[width=\textwidth, trim={1cm, 5.5cm, 1cm, 5.5cm}, clip]{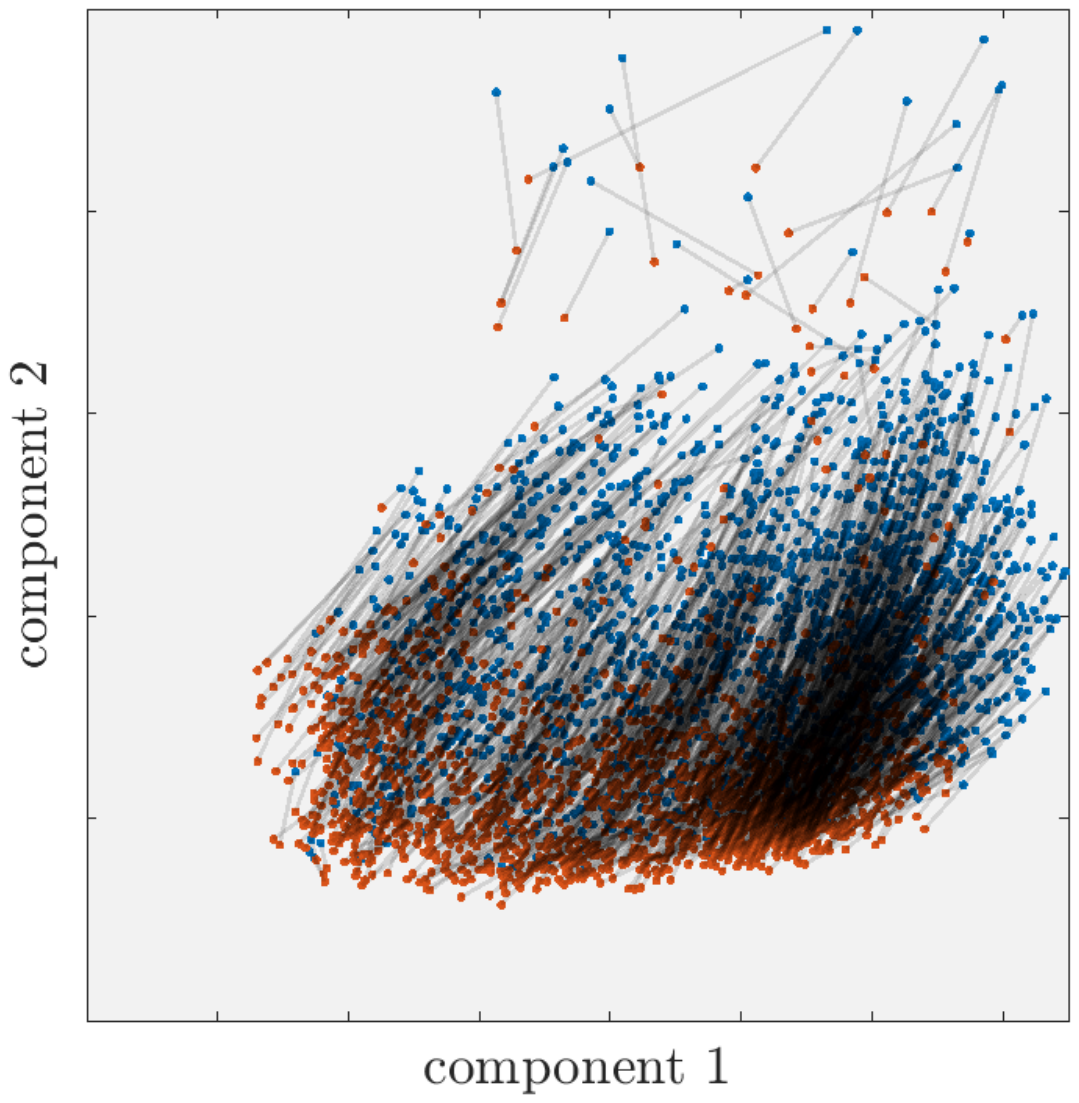}
         \caption{Alignment without constraints (i.e. $\lambda = 0$)}
     \end{subfigure}
    \hfill
      \begin{subfigure}[b]{0.33\textwidth}
         \centering
         \includegraphics[width=\textwidth, trim={1cm, 5.5cm, 1cm, 5.5cm}, clip]{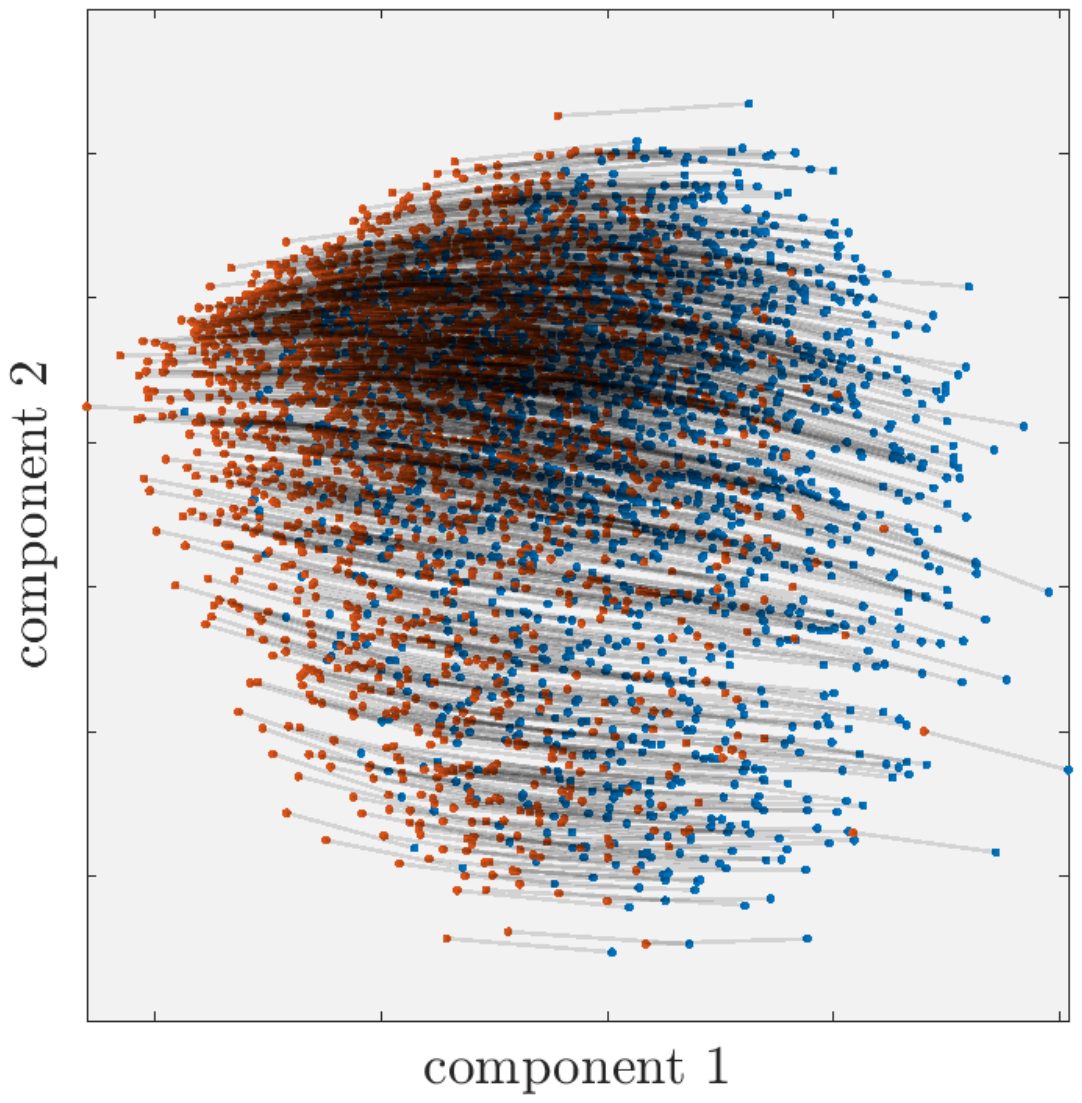}
         \caption{Alignment with constraints (i.e. $\lambda = 1$)}
     \end{subfigure}
\\
     \begin{subfigure}[b]{0.33\textwidth}
         \centering
         \includegraphics[width=\textwidth, trim={1cm, 5.5cm, 1cm, 5.5cm}, clip]{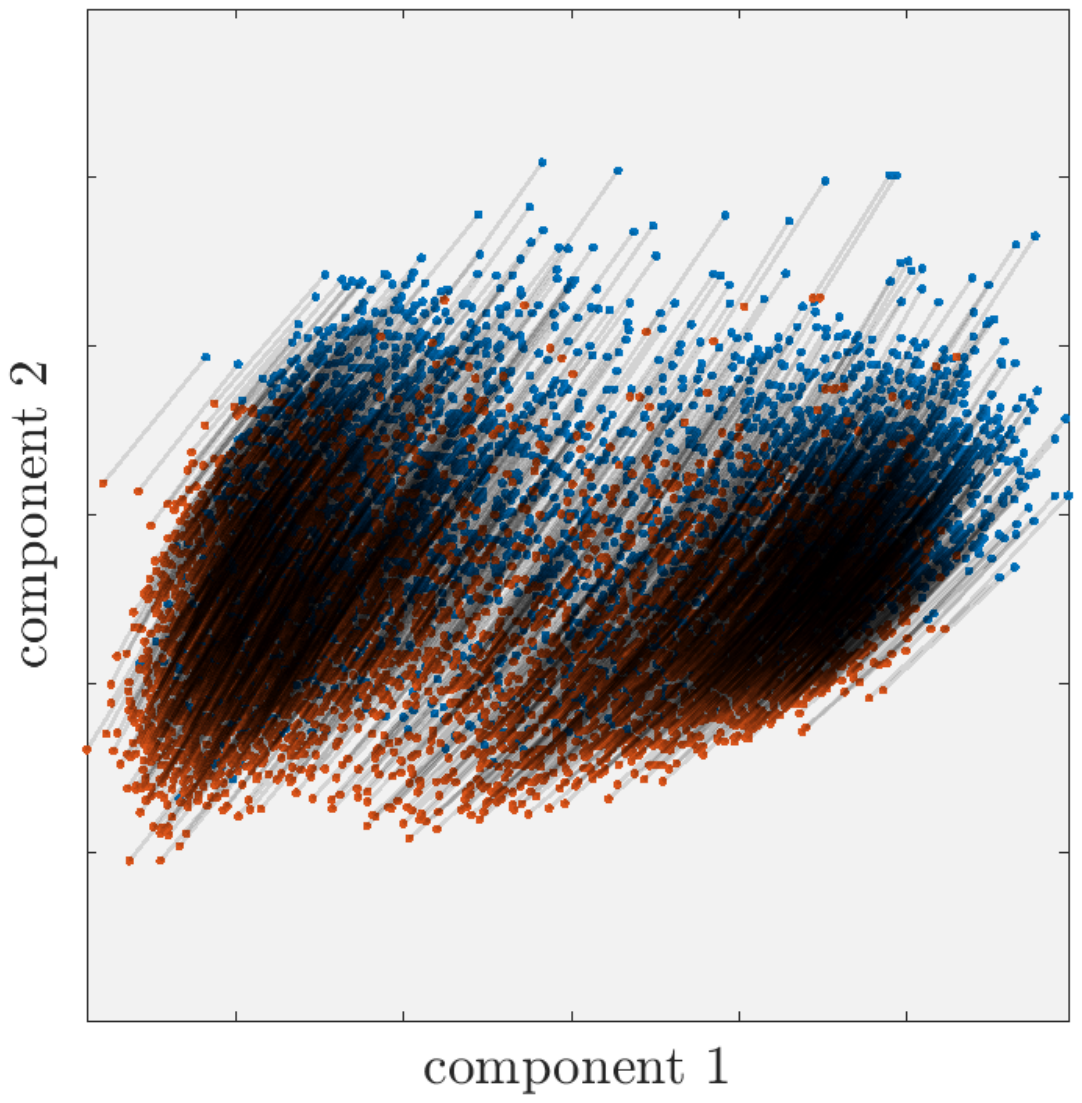}
         \caption{Alignment with Constraints ($\lambda = 1$)}
     \end{subfigure}
    \hfill
     \begin{subfigure}[b]{0.33\textwidth}
         \centering
         \includegraphics[width=\textwidth, trim={1cm, 5.5cm, 1cm, 5.5cm}, clip]{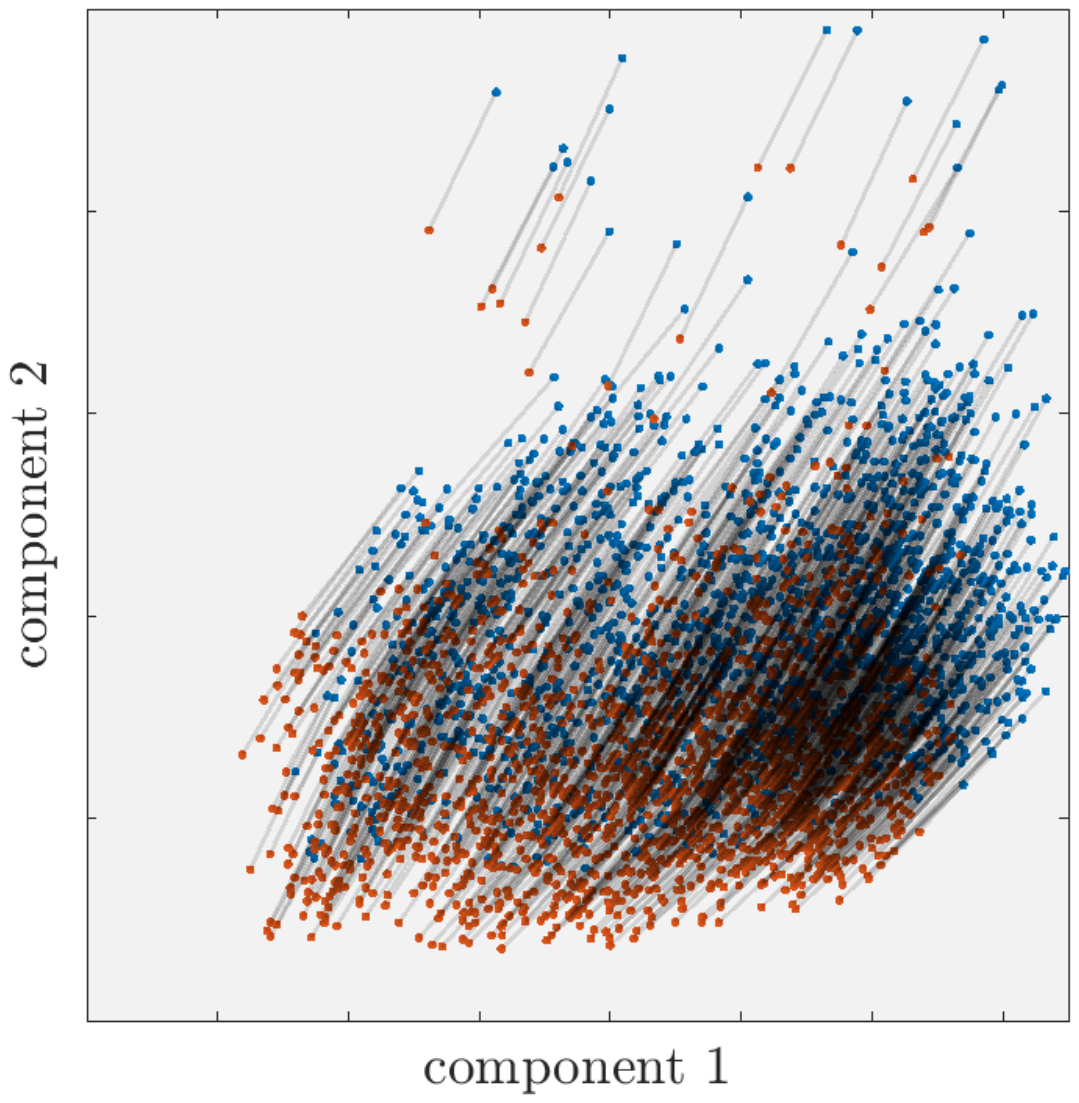}
         \caption{Alignment with constraints (i.e. $\lambda = 1$)}
     \end{subfigure}
    \hfill
     \begin{subfigure}[b]{0.33\textwidth}
         \centering
         \includegraphics[width=\textwidth, trim={1cm, 5.5cm, 1cm, 5.5cm}, clip]{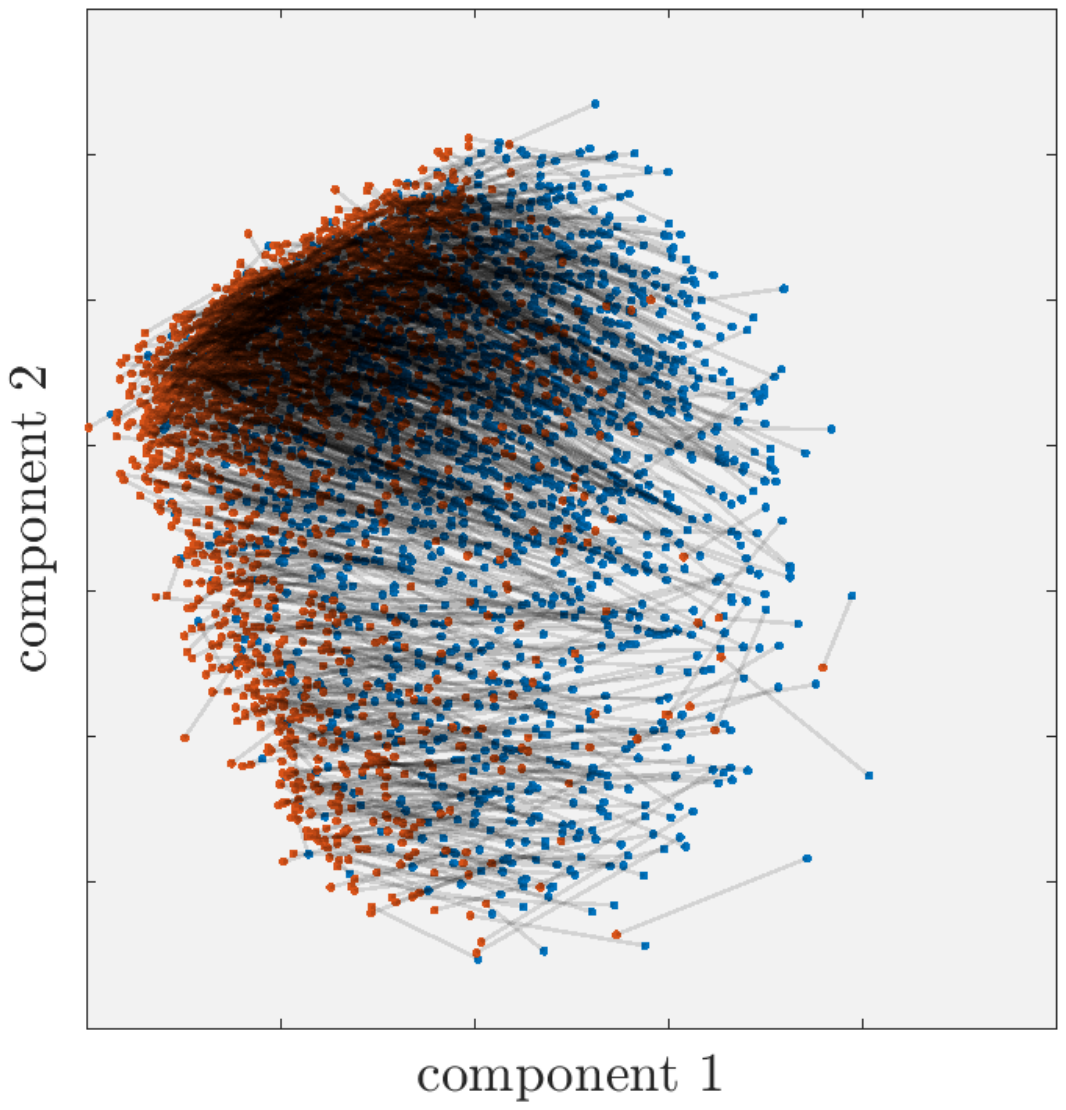}
         \caption{Alignment without constraints (i.e. $\lambda = 0$)}
     \end{subfigure}
    \caption{Point set transformations (alignment) for patient \#2 samples on Day 1 and Day 2, shown in the space of the first two principal components. The arrows indicate the movement of points (cells) during the transformation. (a,b,c) Without topology preservation, cells may move non-coherently. (d,e,f) With topology preservation, cells move coherently, preserving local structures.}
    \label{fig:comparison_person2_warp}
\end{figure}

By adjusting the regularization parameter $\lambda$, we controlled the balance between alignment accuracy and topology preservation. With higher values of $\lambda$, the method prioritized preserving the local topology of the data, maintaining the biological relationships between cells. Figure~\ref{fig:comparison_person2_warp} illustrates the learned transformations with different $\lambda$ values. Although the samples appear less aligned when using a larger $\lambda$, this comes with the benefit of preserving the shape and structure of the original data after transformation, which is desirable in biological settings.

\subsection*{Preservation of Marginal Distributions}

\begin{figure}[h]
\centering
     \begin{subfigure}[b]{0.245\textwidth}
         \centering
         \includegraphics[width=\textwidth, trim={5cm, 4.5cm, 5.5cm, 3.5cm}, clip]{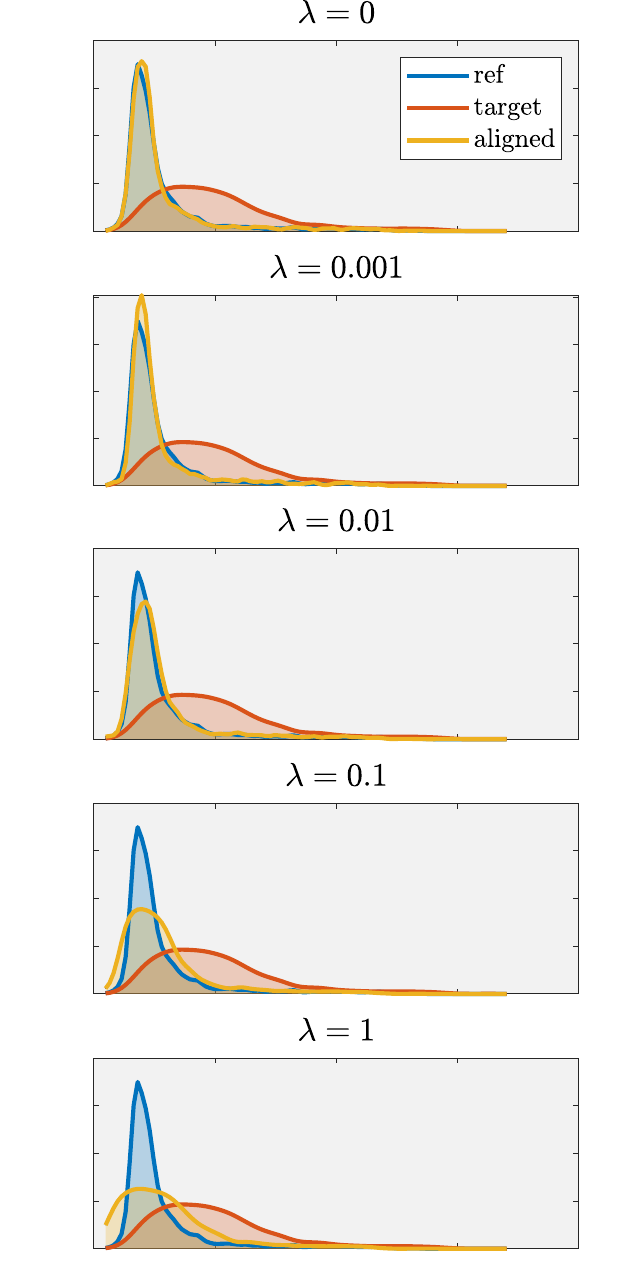}
         \caption{CD28}
     \end{subfigure}
     \begin{subfigure}[b]{0.245\textwidth}
         \centering
         \includegraphics[width=\textwidth, trim={5cm, 4.5cm, 5.5cm, 3.5cm}, clip]{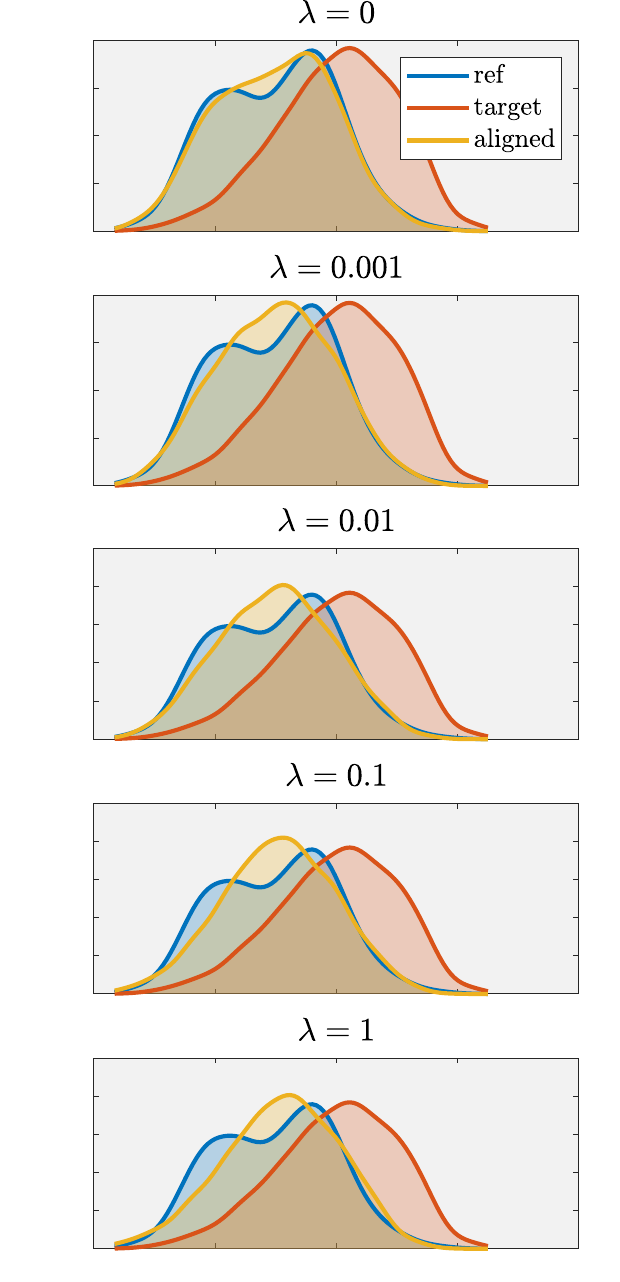}
         \caption{GzB}
     \end{subfigure}
     \begin{subfigure}[b]{0.245\textwidth}
         \centering
         \includegraphics[width=\textwidth, trim={5cm, 4.5cm, 5.5cm, 3.5cm}, clip]{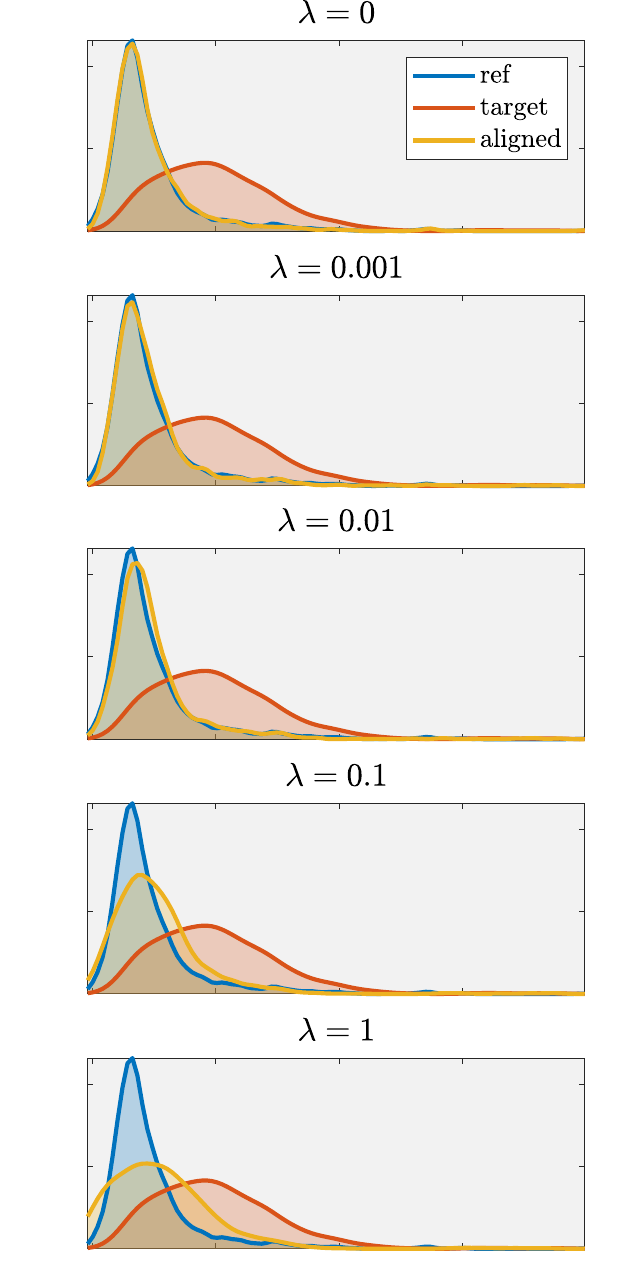}
         \caption{CD19}
     \end{subfigure}
     \begin{subfigure}[b]{0.245\textwidth}
         \centering
         \includegraphics[width=\textwidth, trim={5cm, 4.5cm, 5.5cm, 3.5cm}, clip]{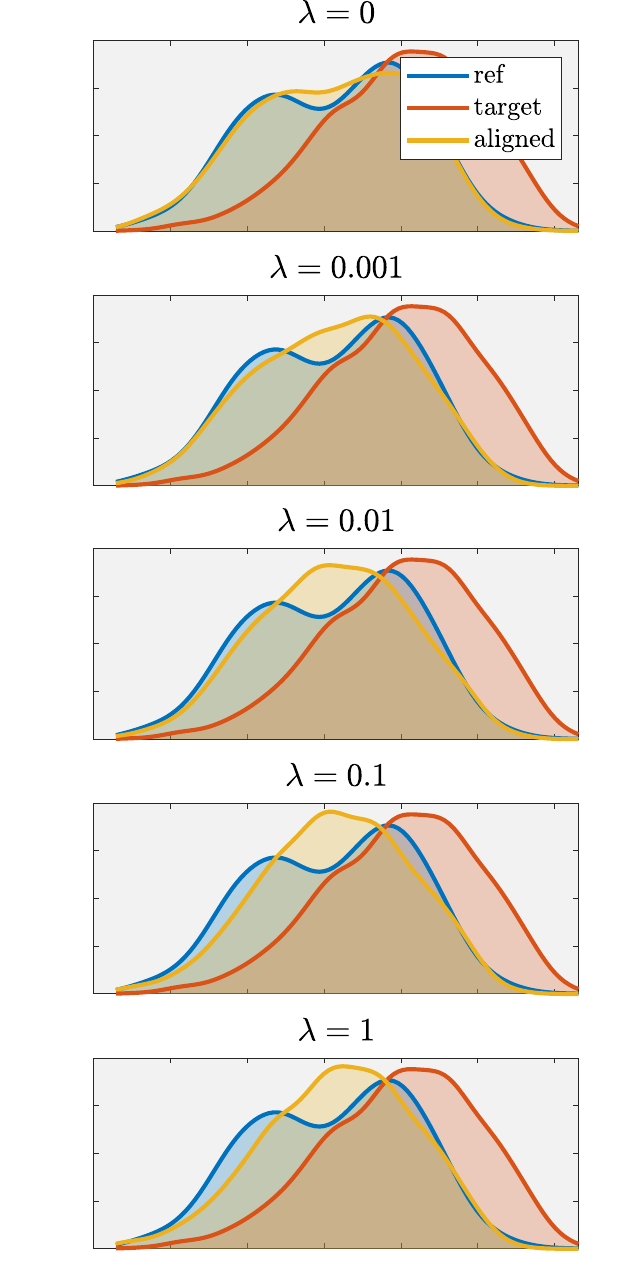}
         \caption{CD45}
     \end{subfigure}
    \caption{Marginal distributions of selected biomarkers for patient \#2 before and after alignment with different $\lambda$ values. The distributions are better preserved with higher $\lambda$, indicating better topology preservation.}
    \label{fig:marginals_comparison}
\end{figure}

Figure~\ref{fig:marginals_comparison} shows the marginal distributions of selected biomarkers before and after alignment for different values of $\lambda$. Our method preserved the marginal distributions of key biomarkers while maintaining the kNN graph structure, leading to more biologically meaningful alignments. It is evident that having a small $\lambda$ favors alignment over faithfulness to the original distribution, whereas increasing $\lambda$ preserves the shape of the original data after transformation.

\subsection*{Comparison with Existing Methods}

We compared our method with traditional batch normalization techniques that do not incorporate topology preservation, such as the method described in~\cite{shaham2017removal}. These methods often align marginal distributions but can distort the local relationships between cells. Our method outperformed these approaches by preserving both the global alignment and the local topological structures, which is crucial for downstream biological analyses.

\section*{Discussion}

Our proposed residual neural network-based method addresses a critical challenge in the analysis of high-dimensional CyTOF data: the effective removal of batch effects while preserving the biological integrity of the data. By aligning cellular distributions across batches and experimental conditions without disrupting the local topological structure, our method facilitates more accurate and meaningful downstream analyses, such as cell population identification, differential expression analysis, and trajectory inference.

In our experiments with real-world CyTOF datasets, we demonstrated that our method effectively aligns samples collected on different days or under different conditions. By preserving the kNN graph structure of the target data, our approach maintains the relationships between cells that are essential for capturing biological phenomena, such as cell differentiation pathways and lineage hierarchies. This is particularly important in biomedical research, where subtle differences in cellular populations can have significant implications for understanding disease mechanisms and developing therapeutic interventions.

Furthermore, our method's ability to handle high-dimensional data through a stochastic approximation of the Jacobian cost function makes it practical for use with modern single-cell datasets, which often include measurements of dozens or even hundreds of markers. This scalability is crucial given the increasing dimensionality of single-cell technologies.

However, our method also has limitations. The reliance on the preservation of the kNN graph assumes that the local neighborhood relationships are biologically meaningful and should be maintained across batches. In cases where batch effects introduce non-linear distortions that significantly alter these relationships, additional strategies may be necessary to accurately capture and correct for such effects. Additionally, the choice of hyperparameters, such as the regularization parameter $\lambda$, can influence the balance between alignment accuracy and topology preservation and may require careful tuning based on the specific characteristics of the data.

\subsection*{Limitations}

It is important to note that the orthogonal Jacobian could be too strong of a condition to preserve the kNN graph: 
\begin{equation}
    (\mathbf{v} - \mathbf{u})^\top \mathbf{J}_\mathbf{u}^\top \mathbf{J}_\mathbf{u} (\mathbf{v} - \mathbf{u}) = (\mathbf{v} - \mathbf{u})^\top (\mathbf{v} - \mathbf{u})
\end{equation}

The objective is satisfied by preserving inequality and not equality. In other words, it is only necessary and sufficient for $\mathbf{J}$ to preserve the kNN graph if the following holds: 
\begin{equation}
   \mathbf{u}^\top \mathbf{u} \leq \mathbf{v}^\top \mathbf{v} \rightarrow \mathbf{u}^\top \mathbf{J}^\top \mathbf{J} \mathbf{u} \leq \mathbf{v}^\top \mathbf{J}^\top \mathbf{J} \mathbf{v} 
\end{equation}
 
or 
\begin{equation}
    \langle \mathbf{u}, \mathbf{u} \rangle \leq \langle \mathbf{v}, \mathbf{v} \rangle \rightarrow \langle \mathbf{Ju}, \mathbf{Ju} \rangle \leq  \langle \mathbf{Jv}, \mathbf{Jv} \rangle
\end{equation}

Having strict equality puts a limitation on the kind of transformations the model is capable of learning. Furthermore, even if the deformation could theoretically be expressed, such a penalty makes convergence unnecessarily slower. 
On the empirical side, we only have a limited number of experiments to test the proposed method. More experimentation and ablation are required to better understand the limits of our current approach and to learn how it fairs on a wider selection of real-world data such as RNA-Seq. 

Although the method demonstrates reduced batch effects and preserved topological structures, its effectiveness in practice requires expert validation. Specifically, domain experts must label and evaluate cells post-transformation to verify that biologically similar cells overlap between the transformed target and reference sets. Without such validation, the biological interpretability of the transformed data cannot be guaranteed.

\subsection*{Future Work}

In future work, our aim is to extend our approach to accommodate partial or local matching, which is common in biological datasets due to the presence of rare cell types or batch-specific populations. Incorporating more sophisticated alignment losses, such as Gromov-Wasserstein distances, could enhance our method's robustness to outliers and missing data. Additionally, applying our framework to other high-dimensional biomedical data types, such as single-cell RNA sequencing or multimodal datasets, could further demonstrate its versatility and impact in the field.

Finally, to enhance the biological relevance of the normalized data, our objective is to collaborate with domain experts to perform manual labeling of cells after transformation. This step will ensure that biologically similar cells align correctly between the transformed target and reference sets, providing a practical assessment of the effectiveness of the model.

\section*{Conclusion}

We have introduced a novel method for batch normalization in high-dimensional CyTOF data that aligns cellular distributions across batches while preserving the local topological structure essential for biological interpretation. Using a residual neural network architecture with Jacobian-based regularization and geometry-aware alignment losses, our approach addresses the limitations of traditional batch normalization methods that can alter biological relationships between cells.

Our method is computationally efficient, scalable to high-dimensional data, and flexible in balancing alignment accuracy with topology preservation through the regularization parameter $\lambda$. Experimental results on real-world CyTOF datasets demonstrate the effectiveness of our approach in reducing batch effects and facilitating reliable comparative analyses.

This work has significant implications for biomedical research, particularly in studies involving single-cell analyses where accurate batch normalization is critical. By preserving the biological integrity of the data, our method enables more accurate identification of cellular populations, understanding of disease mechanisms, and development of therapeutic strategies. Future work may extend our approach to other high-dimensional biological datasets and explore integration with downstream analytical pipelines.

\bibliography{main}

\end{document}